\documentclass[10pt,twocolumn,letterpaper]{article}

\usepackage[pagenumbers]{cvpr} 
\usepackage{multirow}
\usepackage{multicol}
\usepackage{mathtools}
\usepackage{graphicx}
\usepackage{float}
\usepackage{xspace}
\usepackage{bbold}
\usepackage{xcolor}
\usepackage{tabularx}

\definecolor{cvprblue}{rgb}{0.21,0.49,0.74}
\usepackage[pagebackref,breaklinks,colorlinks,citecolor=cvprblue]{hyperref}

\newcommand{\Ours}{MultiDiff\xspace}
\newcommand{\DFM}{DFM\xspace}
\newcommand{\TR}{Text2Room\xspace}
\newcommand{\PhotoNVS}{PhotoNVS\xspace}
\newcommand{\MVD}{MVDiffusion\xspace}
\newcommand{\EPIDIFF}{Pose-Guided Diffusion\xspace}
\newcommand{\VC}{VideoCrafter\xspace}
\newcommand{\SUPP}{supplementary material\xspace}
\def\paperID{7529} 
\def\confName{CVPR}
\def\confYear{2024}

\newcommand{\OURS}{MultiDiff}

\title{\vspace{-30pt}\OURS: Consistent Novel View Synthesis from a Single Image}

\author{
Norman M{\"u}ller$^{1}$~~~
Katja Schwarz$^{1}$~~~ 
Barbara Roessle$^{2}$~~~ 
Lorenzo Porzi$^1$~~~
Samuel Rota Bul\`{o}$^1$~~~\\
Matthias Nie{\ss}ner$^2$~~~
Peter Kontschieder$^1$~~~
\vspace{0.2cm} \\
Meta Reality Labs Zurich$^1\quad$
Technical University of Munich$^2$~~~
\vspace{0.2cm}
}

\begin{document}
\twocolumn[{%
	\renewcommand\twocolumn[1][]{#1}%
	\maketitle
        \vspace{-1cm}
	\begin{center}
            \includegraphics[width=\linewidth, trim={0 0.5cm 0.4cm 0cm},clip]{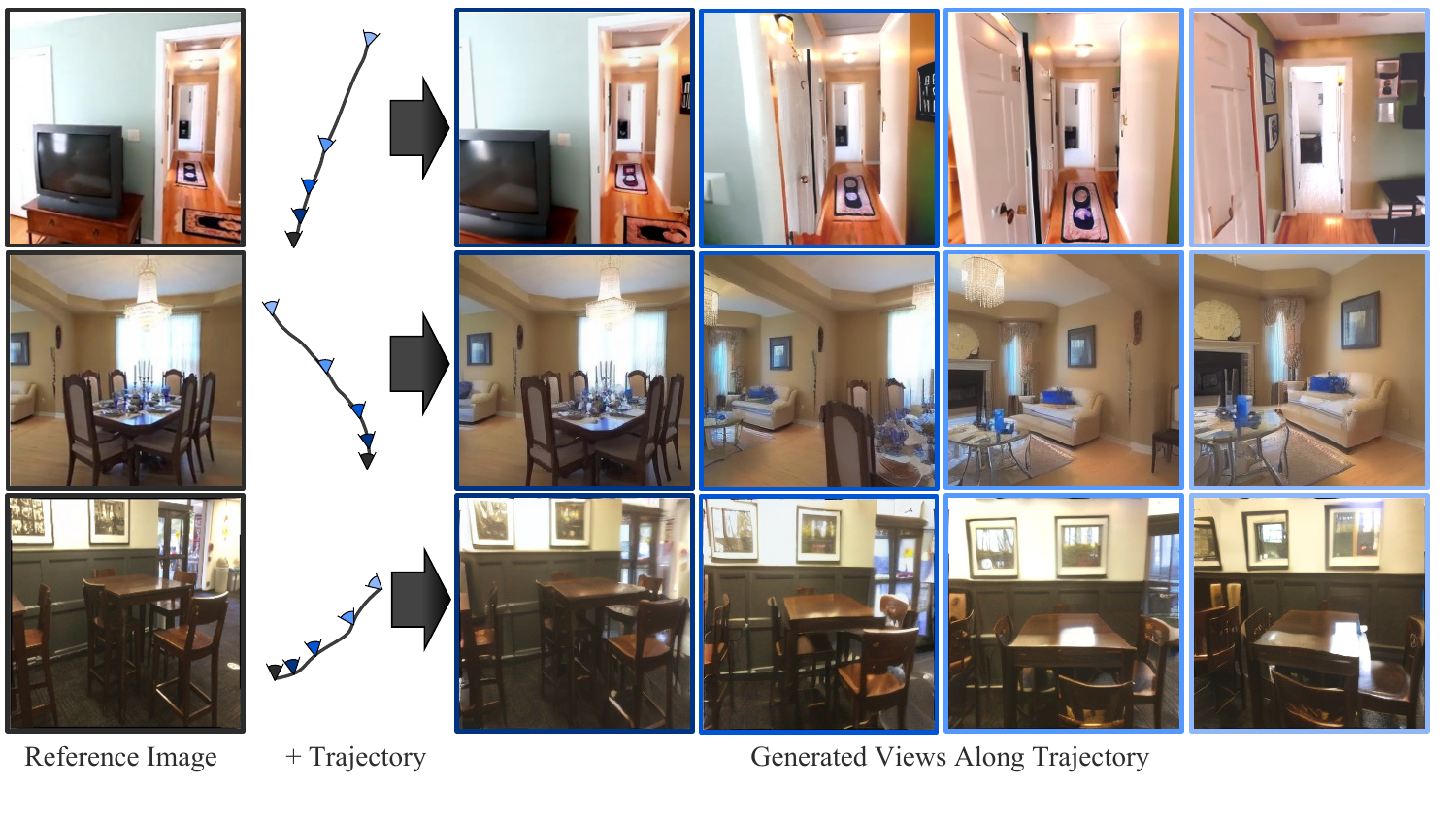}
            \vspace{-2em}
         \captionof{figure}{
         Given a single input image, \Ours synthesizes consistent novel views following a desired camera trajectory. These synthesized views harmonize well even in areas unseen from the reference view.  Examples from RealEstate10K~\cite{realestate46965} (top two rows) and ScanNet~\cite{dai2017scannet} (bottom row) test sets demonstrate that our model can handle large camera changes and challenging perspectives.
            }
		\label{fig:teaser}
	\end{center}    
}]

\begin{abstract}
We introduce \Ours, a novel approach for consistent novel view synthesis of scenes from a single RGB image. 
The task of synthesizing novel views from a single reference image is highly ill-posed by nature, as there exist multiple, plausible explanations for unobserved areas.
To address this issue, we incorporate strong priors in form of monocular depth predictors and video-diffusion models. Monocular depth enables us to condition our model on warped reference images for the target views, increasing geometric stability. The video-diffusion prior provides a strong proxy for 3D scenes, allowing the model to learn continuous and pixel-accurate correspondences across generated images.
In contrast to approaches relying on autoregressive image generation that are prone to drifts and error accumulation, \Ours jointly synthesizes a sequence of frames yielding high-quality and multi-view consistent results -- even for long-term scene generation with large camera movements, while reducing inference time by an order of magnitude.
For additional consistency and image quality improvements, we introduce a novel, structured noise distribution. 
Our experimental results demonstrate that \Ours outperforms state-of-the-art methods on the challenging, real-world datasets RealEstate10K and ScanNet.
Finally, our model naturally supports multi-view consistent editing without the need for further tuning. Project page: \url{https://sirwyver.github.io/MultiDiff/}
\end{abstract}    
\section{Introduction}
\label{sec:intro}

In this work, we address the challenging and highly ill-posed task of view extrapolation from a single image. The goal is to synthesize a set of multiple novel views that are diverse and in themselves consistent. As input, our method only requires a single input image and a user-defined free-form camera trajectory that may deviate substantially from the reference view. Providing a solution to this problem unlocks applications in virtual \& augmented reality and 3D content creation, where generating immersive and multi-view coherent scenes is paramount.

Many existing, state-of-the-art approaches for novel view synthesis are reconstruction-based (\eg, by optimizing a Neural Radiance Field~\cite{mildenhall2020nerf} from a fixed number of input views), and are thus inherently limited in generating high-quality novel views for areas without sufficient training coverage. In contrast, we leverage diffusion-based, generative approaches~\cite{sohl2015deep,song2020denoising,song2020improved,genmodel,ho2020denoising}, that are capable of producing high-quality, single images or individual, simple 3D objects, due to their ability of learning powerful (conditional) image priors. 
Despite significant progress, these models are still unable to synthesize several, multi-view consistent views of large scenes.
This is largely due to the lack of inherent 3D modeling capabilities, the absence of large-scale 3D ground truth datasets, but also the ill-posed nature of the problem, requiring more sophisticated methodological advances. Ultimately, we are aiming for a solution that i) generates seamlessly aligned and multi-view consistent output images \wrt a given input image, ii) maintains both high variability and fidelity in occluded regions and previously unseen areas, and iii) extends to camera trajectories well beyond the provided input reference image viewpoint or a simplistic 360$^\circ$~panoramic view. 

Some recent works have approached consistent view extrapolation by leveraging an autoregressive approach: \textit{Look Outside the Room}~\cite{ren2022look} is a transformer-based approach combined with locality constraints \wrt the input cameras for enforcing consistency among generated frames. Similarly, \textit{Pose-Guided Diffusion Models}~\cite{tseng2023consistent} apply attention along epipolar lines to condition a diffusion model. PhotoNVS~\cite{Yu2023PhotoconsistentNVS} also proposes an autoregressive attempt where the diffusion model is conditioned on a reference view and a specialized representation for relative camera geometry. A significant drawback of autoregressive models is their tendency to error accumulation~\cite{liu2021infinite,pmlr-v15-ross11a}. Repeatedly conditioning the model on its previously generated frames can turn minor output deficiencies quickly into undesirable and semantically meaningless results -- particularly on longer-term trajectories. In contrast, \textit{Diffusion with Forward Models}~\cite{tewari2023diffusion} (DFM) trains a diffusion model to directly sample from the distribution of 3D scenes, inherently improving 3D consistency. However, DFM is computationally expensive, limited to low image resolutions, slow at inference, and cannot directly integrate 2D diffusion priors. The goal of our work is to overcome both the main limitations of autoregressive works and enabling fast and significantly more stable, long-term generation of novel views.

To this end, we propose \Ours, a novel and improved, latent diffusion model-based approach for novel view synthesis, given a single reference image and a pre-defined target camera trajectory as input. We address the challenge of generating pixel-aligned, multi-view consistent image sequences by incorporating strong and complementary priors, significantly constraining the ill-posed nature of the task. Geometric stability is improved by integrating a monocular depth prior, where we condition our model on warped reference images for desired novel views, using off-the-shelf but potentially noisy monocular depth estimators. We also introduce a structured noise distribution for improving multi-view consistency, applying the aforementioned warping procedure to the reference image noise and hence generating correlated 3D noise in all overlapping target views.

By integrating a video diffusion model prior, we are able to compensate for missing and geometrically inconsistent reference image warpings due to potential issues with the monocular depth estimator. Video priors provide a strong proxy for 3D scene understanding, enhancing temporal consistency by largely reducing flickering artifacts -- particularly for long-trajectory view synthesis. However, their lack of explicit camera control makes their integration nontrivial for view extrapolation.

In order to avoid error propagation issues as observed with autoregressive models, we synthesize entire sequences of novel views in a concurrent and efficient way. Finally, due to our conditioning, we can additionally edit our generated scenes, allowing for direct and intuitive interaction with our model.
We summarize our main contributions as follows:
\begin{itemize}
    \item We address the ill-posed view extrapolation problem by integrating priors from monocular depth estimators and video diffusion models for learning pixel-wise correspondences using novel techniques for spatial-aware conditioning across predicted sequences. 
    \item We simultaneously predict multiple frames for a target sequence, overcoming error accumulation of autoregressive methods, while retaining higher resolution at reduced computational costs compared to methods directly sampling from the distribution of 3D scenes.
    \item By introducing a novel structured noise distribution, we obtain more multi-view consistent sampling results.
\end{itemize}

\section{Related Works}
\label{sec:relworks}

\paragraph{Image and Video Diffusion.}
Diffusion Models (DMs)~\cite{ho2020denoising,sohldickstein2015deep,song2020score} are powerful generative models that have achieved state-of-the-art results in unconditional as well as class- and text-guided image synthesis~\cite{nichol2021improved,rombach2021highresolution,dhariwal2021diffusion,ho2022cascaded,dockhorn2022langevin,nichol2022glide,vahdat2021latent,ramesh2022hierarchical,balaji2022ediffi,saharia2022photorealistic,podell2023sdxl,gu2022vector,he2023scalecrafter}. Recently, DMs have been extended to the task of video synthesis~\cite{ho2022imagenvideo,ho2022video,singer2022make,luo2023videofusion,esser2023structure,videocrafter}. 
While recent video DMs can be conditioned on different modalities such as text or images~\cite{videocrafter,wang2023videocomposer,gu2023seer}, they do not enable explicit control the camera viewpoint in the generated videos. Nonetheless, the temporal consistency learnt by these models is a powerful prior that we can leverage to tackle the task of novel view synthesis in an underconstrained setting. Specifically, we use the publicly available VideoCrafter1~\cite{videocrafter} to initialize the correspondence attention layers in our pipeline. 
\vspace{-1em}
\paragraph{Regression-Based Models for Novel View Synthesis.}
The goal of novel view synthesis (NVS) is to produce realistic images of a given instance or scene from previously unseen camera viewpoints. Earlier approaches require hundreds of posed training images per instance and optimize each instance individually~\cite{mildenhall2020nerf,lombardi2019neural,meshry2019wild,dai2020pcrender,thies2019deferred,sitzmann2019deepvoxels, mueller2022autorf}. By learning priors across multiple training scenes, more recent works enable NVS from only one or a few images at inference ~\cite{yu2021pixelnerf,sitzmann2019scene,niemeyer2022,niemeyer2020dvr,wang2021ibrnet,Henzler_2021_CVPR,du2023wide,chen2021mvsnerf,trevithick2020GRF,sajjadi2022scene,kulhanek2022viewformer,roessle2022depthpriorsnerf}. These methods optimize a regression objective, i.e. an L1 or L2 loss to reconstruct the training images. While this allows for impressive results on interpolation near input views, regression-based NVS approaches struggle with reconstruction ambiguity and longer-range extrapolations~\cite{chan2023genvs}. As our goal is to synthesize novel views far beyond observed views, we instead train a generative model. 
\vspace{-1em}
\paragraph{Generative Models for Novel View Synthesis.}
To better model reconstruction ambiguity and long-range view extrapolation, multiple recent works deploy generative models for NVS. Earlier works use GANs~\cite{wiles2020synsin,koh2022simple,li2022infinite,pavllo2023shape,koh2021pathdreamer}, VAEs~\cite{liu2021infinite}, or autoregressive models~\cite{rockwell2021pixelsynth,rombach2021geometry,ren2022look}. Interestingly, GeoGPT~\cite{rombach2021geometry} directly models long-range 3D correspondences between source and target views with an autoregressive transformer, demonstrating that an intermediate 3D representation may not be needed for NVS from a single image. 
More recently, diffusion models have achieved impressive results on object-centric data~\cite{xiang2023ivid,watson20233dim,sargent2023zeronvs,liu2023zero1to3,zhou2023sparsefusion,anciukevicius2022renderdiffusion, muller2023diffrf}. While these works focus on relatively constrained camera motions around a single object, another line of work addresses scenes with arguably more complex camera trajectories~\cite{bautista2022gaudi,tang2023MVDiffusion,kim2023nfldm,hoellein2023text2room,fridman2023SceneScape,chan2023genvs,tewari2023diffusion,tseng2023consistent,Yu2023PhotoconsistentNVS}. 
\MVD~\cite{tang2023MVDiffusion} performs image synthesis conditioned on depth maps of a given mesh, jointly generating all images of the trajectory. To increase consistency, cross-view interactions are modelled by correspondence-aware attention layers that require given pixel-to-pixel correspondences using GT geometry during training and inference. 
Our approach instead aims at learning those multi-view correspondences, which allows us to synthesize novel views along a trajectory given just a single RGB image, without the need for any geometric information about the target views. This renders our method applicable to a wider range of scenarios, where no prior 3D reconstruction is available.
\DFM~\cite{tewari2023diffusion} trains a diffusion model to directly sample from the distribution of 3D scenes. Modeling the scene with a 3D representation is inherently 3D-consistent, but is computationally expensive and in practice limits \DFM to lower image resolutions and slow inference. \EPIDIFF~\cite{tseng2023consistent} and \PhotoNVS~\cite{Yu2023PhotoconsistentNVS} train a pose-conditioned 2D diffusion model to autoregressively generate frames along a given camera trajectory. However, especially for long trajectories, autoregressive sampling is prone to error accumulation, leading to common struggles with loop closure when taking a Markov assumption and slow inference as it cannot be parallelized. 
Hence, we do not use autoregressive sampling but generate all images jointly, enabling the model to learn short- and long-term correspondences between views. In stark contrast to \MVD that also performs joint frame synthesis, we only use depth from an off-the-shelf monocular depth estimator with no geometric cues about the target views. \Ours can therefore generate novel views from a single input image only. The learnt correspondence attention enables our model to achieve better consistency than state-of-the-art autoregressive approaches while achieving higher image quality than related works. 
 
\section{Method}
\label{sec:method}

\begin{figure*}
\begin{center}
\includegraphics[width=.98\textwidth]{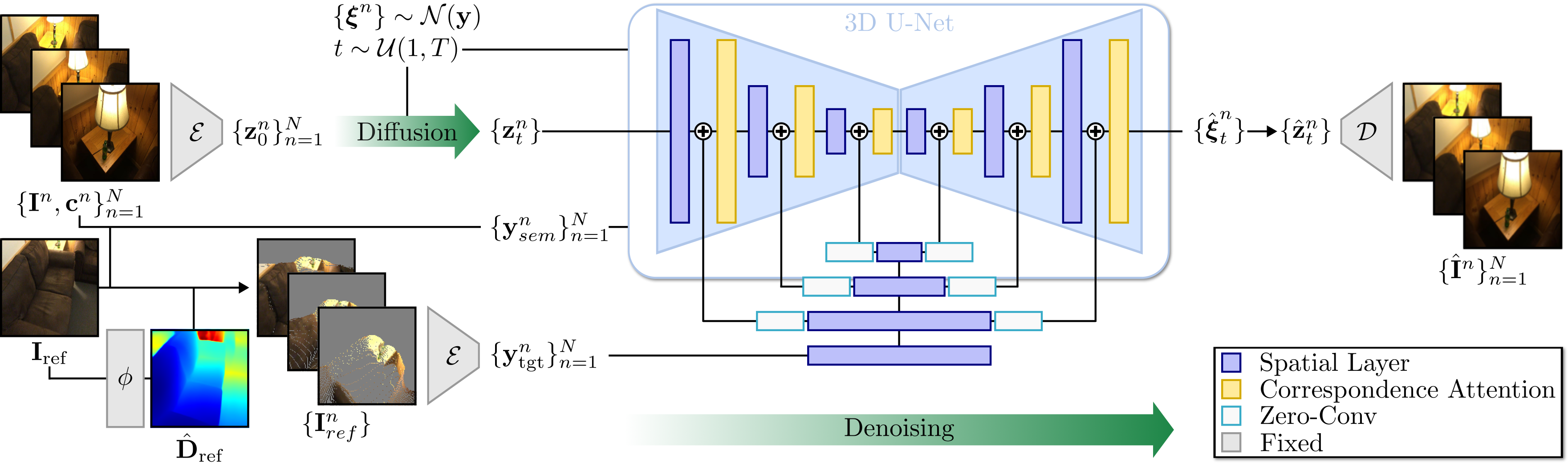}

\end{center}
\vspace{-1em}
\caption{
\Ours is pose-conditional diffusion model for novel view synthesis from a single image. The diffusion model is trained in the latent space of a fixed auto-encoder with encoder $\mathcal{E}$ and decoder $\mathcal{D}$ and is conditioned on a reference image $\mathbf{I}_\text{ref}$ and a camera trajectory $\{\mathbf{c}^n\}$. 
Specifically, we embed $N$ posed target images $\{\mathbf{I}^n\}_{n=1}^N$ into latent space, apply forward diffusion according to a timestep $t$ and structured noise $\{\boldsymbol{\xi}^n\}$, and train a 3D U-Net to predict $\{\boldsymbol{\xi}^n\}$ from the noisy inputs $\{\mathbf{z}_t^n\}$. For each sample $n$, the U-Net’s prediction $\hat{\boldsymbol{\xi}}_t^n$ is used to reconstruct the denoised sample $\hat{\mathbf{z}}_t^n$ which can then be decoded into the predicted target image $\hat{\mathbf{I}}^n$. 
We condition the U-Net on the reference image by warping $\mathbf{I}_\text{ref}$ to the novel views using depth $\hat{\mathbf{D}}_\text{ref}$ from a pretrained estimator $\phi$. The warps $\{\mathbf{I}_\text{ref}^n\}$ are encoded into latent representations $\{\mathbf{y}_\text{tgt}^n\}$ and injected into the U-Net in a ControlNet inspired manner. We further condition the model directly on the camera pose and an embedding of the reference image as part of the semantic condition $\{\mathbf{y}_{sem}^n$\}.
}
\label{fig:method}
\vspace{-5mm}
\end{figure*}

Given a single \emph{reference} image $\mathbf{I}_\text{ref}$, our goal is to generate semantically plausible, consistent novel views along a camera trajectory $\mathcal C\coloneqq \{\mathbf{c}^n\}_{n=1}^N$, where each camera pose $\mathbf c^n$ is relative to the camera of the reference image. 
To this end, we propose a pose-conditional 2D diffusion model with correspondence attention, \ie, attention layers that \emph{jointly} operate on all generated views of the trajectory.
A key challenge in novel-view synthesis for the highly under-constrained single-image setting is to achieve consistency in the lack of explicit correspondence supervision. 
We therefore leverage strong priors that excel at related tasks. Most importantly, we note that the task of video generation is closely related to our problem setting, where temporal consistency is an intrinsic objective. 

In the following, we explain how we can integrate and adjust a video prior in conjunction with depth and image priors to enable free viewpoint control. 
Next, we provide a detailed explanation of our conditioning mechanism and the correspondence attention which adds viewpoint control to our pipeline. 
Lastly, we introduce structured noise, which ports approximate correspondences between frames to obtain more consistent synthesis results. Our pipeline is illustrated in Figure~\ref{fig:method}. 
\vspace{-1em}
\paragraph{Video Prior} \label{sec:video_prior}
We build our generative model on top of \VC~\cite{videocrafter}. 
\VC trains a denoising 3D U-Net in a fixed latent space, using a pretrained image encoder $\mathcal{E}$ and a pretrained image decoder $\mathcal{D}$ to map to and from latent space, respectively.
At the core of \VC is a 3D U-Net with alternating spatial layers and temporal attention. The spatial layers process each frame in a batch individually while the temporal attention operates on all frames jointly.
This pretrained 3D U-Net architecture is a well-suited initialization for the task of NVS as the temporal layers already provide an inductive bias towards (temporal) consistency. 
During training, we nevertheless finetune all layers of the U-Net for the novel view synthesis task, where instead of ensuring temporal consistency, the attention layers should establish correspondences between multiple views. Hence, we refer to this type of attention as \emph{correspondence attention}.
\vspace{-1em}
\paragraph{Novel-view synthesis} 
In order to generate novel views that adhere to the given camera trajectory $\mathcal C$, we need to condition our pipeline on the target camera poses $\mathbf{c}^n\in\mathcal C$. A na\"ive approach to integrating control over the camera viewpoint is to directly condition the 3D U-Net on $\mathbf{c}^n$, e.g., via cross-attention. In practice, we concatenate them to the semantic condition of \VC that consists of an embedding of the reference image and the framerate of the input sequence yielding the semantic conditioning $\mathbf{y}_{sem}$.

However, this form of guidance alone is too weak to deliver satisfactory novel-view synthesis results (see \cref{sec:ablations}). 
We therefore integrate a monocular depth prior in order to constrain the highly ill-posed nature of the task. In our experiments, we use ZoeDepth~\cite{ZoeDepth} pretrained on ScanNet ~\cite{dai2017scannet} and refer to the supplementary material for ablations about alternative monocular estimators. 
We use the depth $\mathbf{D}_\text{ref}$ estimated from the reference image $\mathbf I_\text{ref}$ to implement a warping function $\Psi^n$ that enables warping images from the camera of the reference image to any other camera $\mathbf c^n\in \mathcal C$.
We denote by $\mathbf I_\text{ref}^n\coloneqq\Psi_n(\mathbf{I}_\text{ref})$ the reference image warped to camera $\mathbf c^n$ and by $\mathbf M^n\coloneqq\Psi_n(\mathbf 1)$ the mask indicating the area of valid warped pixels in camera $\mathbf c^n$.
To facilitate learning the 3D correspondences across the spatial features, for each view $n$, we encode $\mathbf I_\text{ref}^n$ into latent space via $\mathcal E$ and stack the mask $\mathbf{M}^n$, suitably resized, along the channel dimension. The resulting tensor is denoted $\mathbf y_\text{tgt}.$
Inspired by ControlNet~\cite{zhang2023controlnet}, we create a copy of the downsampling layers of diffusion U-Net to extract features from  $\mathbf y_\text{tgt}$, but we prepend a convolutional layer to cope with the additional mask channel.

The intermediate feature maps are then processed with zero-initialized convolutions and added to the outputs of all spatial layers of the 3D U-Net. Note that this differs from the procedure proposed in ControlNet, which only inserts the feature maps into the decoder. We further do not freeze the layers of \VC to enable learning the correspondence attention. In initial experiments we found that finetuning all layers jointly results in better performance than using a fixed video prior. 

The warping operation is implemented by leveraging an off-the-shelf monocular depth estimator and thus error-prone and incomplete. 
By also passing the reference image and camera poses to the network in the semantic conditioning $\mathbf{y}_{sem}$, 
we enable our approach to follow the provided trajectory even in absence of overlap with the reference image. We refer to our ablation ~\cref{sec:ablations} for a discussion about the importance of the individual design decisions. 
In the rest of the section we summarize with $\mathbf y$ all quantities we condition our model on, namely reference image $\mathbf{I}_\text{ref}$, camera trajectories $\mathcal C$, and all derived ones (estimated depth, warped reference images, corresponding masks).
\vspace{-1em}
\paragraph{Structured noise distribution $\mathcal N(\mathbf y)$.} \label{sec:struct_noise} Images of a 3D scene captured from different point of views exhibit strong correlations. Hence, it is beneficial to inject similar correlations in the noise $\boldsymbol\epsilon$ that is used by our diffusion model to synthesize the different camera views, which would otherwise be a standard normal multi-variate. This helps enforcing more consistent outputs~\cite{qiu2023freenoise}. Since the correlations are mainly driven by geometric constraints, we leverage the warping function $\Psi_n$ introduced in the previous paragraph to warp a standard normal multi-variate  $\boldsymbol\epsilon^0$ to all other camera views in $\mathcal C$, while filling the gaps with independent Gaussian noise. This yields per-view noise $\boldsymbol\xi^n\coloneqq \mathbf M^n\odot\Psi_n(\boldsymbol\epsilon^0)+(1-\mathbf M^n)\odot\boldsymbol\epsilon^n$, where $\boldsymbol\epsilon^n$ is a standard normal multi-variate and $\mathbf M^n$ is the suitably-resized warp-validity mask. This process yields $\boldsymbol\xi\coloneqq(\boldsymbol\xi^1,\ldots,\boldsymbol\xi^N)$, which is regarded as a sample of the structured noise distribution we denoted by $\mathcal N(\mathbf y)$.

\vspace{-1em}
\paragraph{Training Objective.}
Let $V\coloneqq\{(\mathbf{I}^0,\mathbf c^0),\ldots,(\mathbf{I}^N,\mathbf c^N)\}$ be a ground-truth, posed video sequence, where $\mathbf{I}^n$ and $\mathbf c^n$ are the $n$th image and camera pose in the sequence, respectively.
We assume $\mathbf{I}^0$ to be the reference image, \ie $\mathbf{I}_\text{ref}\coloneqq \mathbf{I}^0$, and assume all cameras to be relative to $\mathbf c^0$. 
We encode all target images of the sequence into a joint latent representation $\mathbf z\coloneqq (\mathbf z^1,\ldots, \mathbf z^N)$, where $\mathbf z^n\coloneqq \mathcal E(\mathbf{I}^n)$, and $\mathbf y$ is the conditioning information encompassing the encoded reference image, camera poses and warped reference images described earlier in the section. The denoising training objective takes the following form for the training example $V$:
\begin{equation}
\mathcal L(\theta;V)\coloneqq \mathbb E_{\substack{\boldsymbol\xi\sim\mathcal N(\mathbf y)\\t\sim\mathcal U(1,T)}}\left [\left \Vert\boldsymbol\xi-\varepsilon_\theta(\mathbf z\oplus_t\boldsymbol\xi;\mathbf y, t) \right\Vert^2\right]\,,
\end{equation}
where $t$ is sampled from a uniform distribution $\mathcal U(1,T)$ and $\boldsymbol\xi$ is noise sampled from the structured noise distribution $\mathcal N(\mathbf y)$. 
The term $\mathbf z \oplus_t \boldsymbol\xi\coloneqq \sqrt{\alpha_t} \mathbf z+\sqrt{1-\alpha_t}\boldsymbol\xi$ perturbs $\mathbf z$ with noise $\boldsymbol\xi$ according to a variance-preserving formulation with parameters $\alpha_t$, from which $\varepsilon_\theta$ , \ie our denoising 3D U-Net with weights $\theta$, is required to recover $\boldsymbol\xi$.

Our model is optimized using Adam by minimizing the training loss function averaged over random batches of video sequences sampled from a given dataset. 
\vspace{-1em}
\paragraph{Inference.}
At inference time, we assume to be given a reference image $\mathbf I_\text{ref}$ and a sequence of cameras $\mathcal C$ relative to it, which we use to compute the conditions $\mathbf{y}$. We generate a video sequence from our model by using the DDIM schedule~\cite{song2020denoising}, \ie starting from $\mathbf z_T\sim \mathcal N(\mathbf y)$ we iterate the following equation
\begin{equation}
\begin{aligned}
\mathbf z_{t-1}\coloneqq &\sqrt{\alpha_{t-1}}(\mathbf z_t\ominus_t \varepsilon_\theta(\mathbf z_t; \mathbf y, t))\\&+\sqrt{1-\alpha_{t-1}}\varepsilon_\theta(\mathbf z_t; \mathbf y, t)\,,
\end{aligned}
\end{equation}
until we obtain $\mathbf z_0$ by setting $\alpha_0\coloneqq 1$.
The term $\mathbf z_t\!\ominus_t\!\boldsymbol\epsilon\coloneqq\frac{\mathbf z_t-\sqrt{1-\alpha_t}\boldsymbol\epsilon}{\sqrt{\alpha_t}}$ recovers $\mathbf z$ from $\mathbf z_t$ assuming noise $\boldsymbol\epsilon$.
The final result $\mathbf z_0$ entails the synthesized views in latent space for all cameras in $\mathcal C$, from which we compute the counterparts in pixel space by applying the decoder $\mathcal D$.
Note that \Ours can generate all images of the sequence simultaneously. However, sometimes the novel view has little or no overlap with the reference image, making the warped reference image, i.e. condition $\mathbf{y}_\text{tgt}^n$ less informative. 
To further refine the results, we can run the sampling again on the generated sequence, but now use the warp of the cloest generated image in $\mathbf{y}_\text{tgt}^n$ which in practice this slightly improves consistency.
\section{Experiments}
\label{sec:experiments}
In this section, we evaluate the performance of our method for the task of consistent novel view synthesis from a single reference image.

\begin{figure*}
\begin{center}
\includegraphics[width=.99\textwidth, trim={0.3cm 0cm 2.7cm 0cm},clip, page=4]{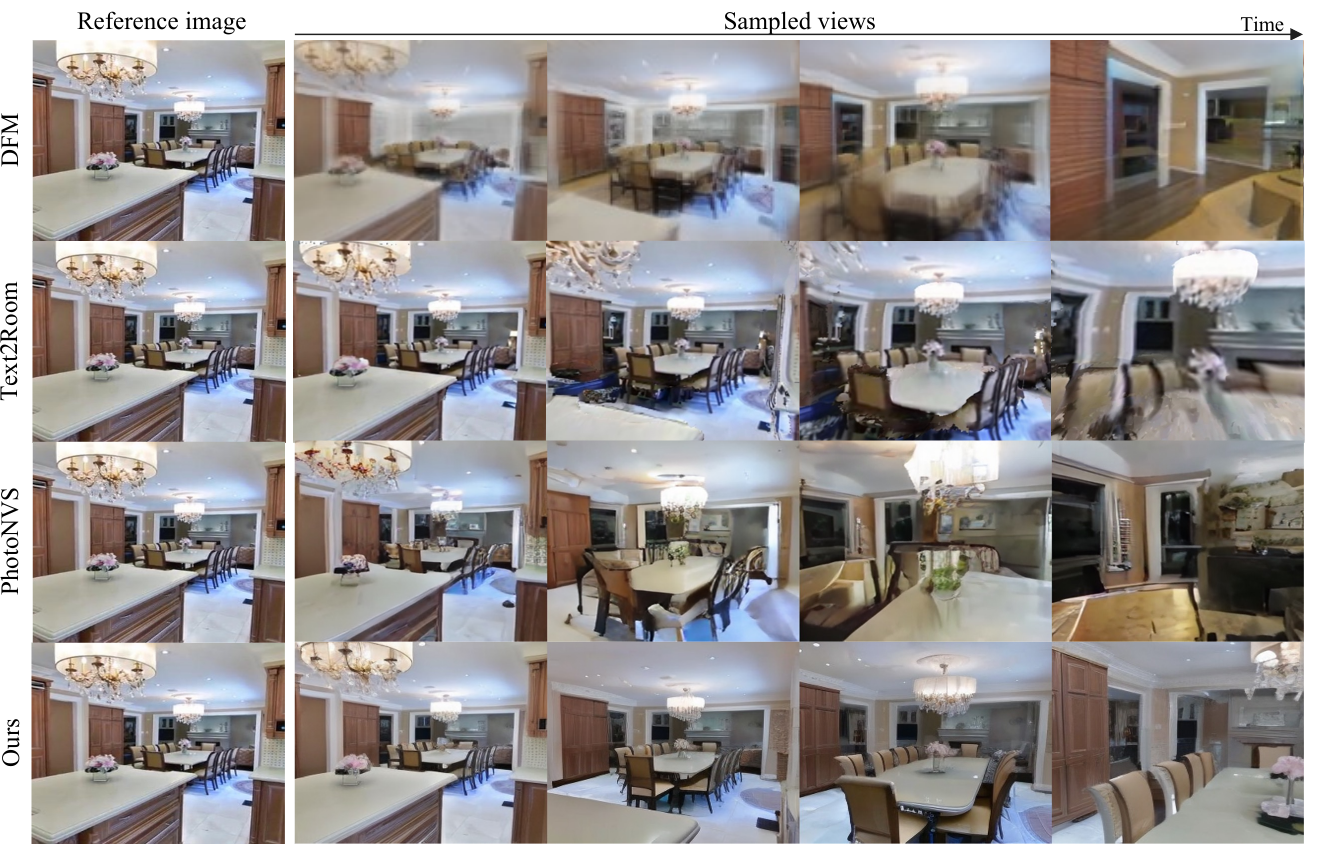}
\end{center}
\vspace{-1em}
\caption{
Novel views following ground-truth trajectories (right) given a reference view (left) on RealEstate10K. Through our joint multi-frame prediction combined with effective priors and conditioning, our sequence of novel views is highly realistic and view-consistent compared to the baselines, which show severe degradation over time.
}
\label{fig:re10k_results}
\vspace{-5mm}
\end{figure*}

\vspace{-1em}
\paragraph{Datasets}
\label{sec:datasets}
We compare our methods against state-of-the-art approaches on RealEstate10K~\cite{realestate46965} and ScanNet~\cite{dai2017scannet}. Both datasets provide video sequences together with registered camera parameters. RealEstate10K is a large dataset of real estate recordings gathered from YouTube. The clips typically feature smooth camera movement with little to no camera roll or pitch. Most frames further show considerable coverage of the respective rooms. Following previous works \cite{ren2022look, Yu2023PhotoconsistentNVS}, we center-crop and downsample the videos to 256px resolution.
ScanNet consists of 1513 hand-held captures of indoor environments. The camera trajectories follow a scan-pattern which can contain rapid changes and variation of camera orientation. The resulting frames encompass close-up object captures as well as wide room recordings, leading to heavy occlusions and an overall diverse data distribution. 
The aforementioned features make ScanNet extremely challenging for novel view synthesis from a single image and our evaluations in~\cref{sec:evaluations} indicate that additional priors are very beneficial in this setting. We resize the images to $256 \times 256$ and remark that ScanNet contains 3D meshes that we use for \MVD as it requires predefined correspondences between frames. 

\paragraph{Evaluations} \label{sec:evaluations}
\begin{table}
    \centering
    \resizebox{.99\columnwidth}{!}{
    \begin{tabular}{c|c|cccc|cccc}
          & \multirow{2}{*}{Method} & \multicolumn{4}{c|}{Short-term} & \multicolumn{4}{c}{Long-term} \\
          & & PSNR $\uparrow$ & LPIPS ↓  & FID ↓  & KID ↓  & FID ↓  & KID ↓ & FVD  ↓  & mTSED  $\uparrow$  \\
         \hline
         \multirow{4}{*}{\rotatebox[origin=c]{90}{\resizebox{0.12\linewidth}{!}{128px}}}          & \DFM~\cite{tewari2023diffusion} & \textbf{18.10} & \textbf{0.299} & 36.37 & 0.010 & 31.20 & 0.007 & 120.2 & \textbf{0.972}  \\
         & \TR~[27] & 15.45 & 0.370 & 34.41  & 0.008  & 84.10  &  0.048  & 163.4 & 0.932  \\
         & \PhotoNVS~\cite{Yu2023PhotoconsistentNVS} & 15.66 & 0.376 & 26.39 & 0.006 & 42.99 & 0.016 &  117.5 &  0.907 \\
         & \Ours (Ours) w/o SN & 16.21 & 0.335 & 27.26 & 0.005 & 30.28 & 0.006 & 107.5& 0.936 \\
         & \Ours (Ours)& 16.41  & 0.318  & \textbf{25.30} & \textbf{0.003} & \textbf{28.25} & \textbf{0.004}  & \textbf{94.37} & 0.941 \\
         \midrule
         \multirow{4}{*}{\rotatebox[origin=c]{90}{\resizebox{0.12\linewidth}{!}{256px}}} 
         & \TR~[27] & 14.88 & 0.458 & 35.41  & 0.009  & 91.92  &  0.050  & 178.6 & 0.837  \\
         & \PhotoNVS~\cite{Yu2023PhotoconsistentNVS} & 15.01 & 0.452 & 26.75 & 0.005 & 45.08 & 0.017 & 130.4   & 0.801 \\
         & \Ours (Ours) w/o SN & 15.55 & 0.412 & 29.49 & 0.008 & 33.71 & 0.010 & 116.4  & 0.849  \\
         & \Ours (Ours)& \textbf{15.65} & \textbf{0.393} & \textbf{25.90} & \textbf{0.004} & \textbf{30.15} & \textbf{0.006} &  \textbf{105.9} &  \textbf{0.855} \\

    \end{tabular}}
    \caption{Quantitative comparison on RealEstate10K~\cite{realestate46965} test sequences. Our model achieves higher image quality than state-of-the-art baselines and comparable consistency compared to \DFM. } 
    \label{tab:re10k}
     \vspace{-2mm}

\end{table}

\begin{table}
    \centering
    \resizebox{.99\columnwidth}{!}{
    \begin{tabular}{c|c|cccc|cccc}
          & \multirow{2}{*}{Method} & \multicolumn{4}{c|}{Short-term} & \multicolumn{4}{c}{Long-term} \\
          & & PSNR $\uparrow$ & LPIPS ↓ & FID ↓  & KID ↓ & FID ↓  & KID ↓ & FVD  ↓ & mTSED  $\uparrow$  \\
         \hline
         \multirow{5}{*}{\rotatebox[origin=c]{90}{\resizebox{0.12\linewidth}{!}{128px}}} 

         & \MVD~\cite{tang2023MVDiffusion} & 13.14 & 0.439 & 43.28  & 0.013  & 43.58  &  0.013  & 186.6 & 0.506  \\
         & \DFM~\cite{tewari2023diffusion} & \textbf{16.59} & 0.444 & 75.19 & 0.036 & 111.9 & 0.069 & 167.2 & \textbf{0.912} \\
         & \TR~[27] & 15.01 & 0.452 &  39.87  & 0.008  & 82.44  &  0.0041  & 173.1 & 0.812  \\
         & \PhotoNVS~\cite{Yu2023PhotoconsistentNVS} & 15.23 & 0.440 & 49.19 & 0.019 & 75.23 & 0.038 & 89.04 & 0.479  \\
         & \Ours (Ours) w/o SN & 15.29  &0.372 & 40.36  & 0.008 & 43.61 & 0.011 & 80.71 &  0.752  \\
         & \Ours (Ours) & 15.50  & \textbf{0.356} & \textbf{38.44}  & \textbf{0.007} & \textbf{42.41} & \textbf{0.010} & \textbf{74.10} & 0.776  \\
         \midrule
         \multirow{4}{*}{\rotatebox[origin=c]{90}{\resizebox{0.12\linewidth}{!}{256px}}} 
         & \MVD~\cite{tang2023MVDiffusion} & 12.88 & 0.502 &  50.18 & 0.017 &  51.60 &  0.018  & 230.1  & 0.361   \\
         & \TR~[27] & 14.32 & 0.514 &  46.69  & 0.014  & 93.09  &  0.058  & 201.1 & \textbf{0.631}  \\
         & \PhotoNVS~\cite{Yu2023PhotoconsistentNVS} & 14.61 & 0.542 & 63.21 & 0.033 & 96.85 & 0.059 & 134.2 & 0.263    \\
          & \Ours (Ours) w/o SN & 14.80 & 0.445 & 47.10 & 0.013 & 50.84 & 0.016 &  119.3  &    0.529 \\
         & \Ours (Ours) & \textbf{15.00} & \textbf{0.431} & \textbf{43.84}  & \textbf{0.010}  & \textbf{47.11}  &  \textbf{0.013}  & \textbf{114.9}  &  0.576  \\

    \end{tabular}}
    \caption{Quantitative comparison on ScanNet~\cite{dai2017scannet} test sequences. Our approach outperform all baselines at 256px resolution and shows significantly higher image fidelity compared to \DFM.} 
    \label{tab:scannet}
      \vspace{-4mm}
\end{table}

We evaluate our approach in terms of image fidelity and consistency of the generated outputs. 
Similar to~\cite{ren2022look}, we consider both short-term and long-term view synthesis. Specifically, we randomly select 1k sequences with 200 frames from the test set and evaluate the 50th generated frame for short-term and the 200th generated frame for long-term view synthesis for RealEstate10K. Due to the faster camera motion, on ScanNet instead we choose the 25th frame for short-term and 100th for long-term evaluation. 
In the short-term setting, we report Peak Signal-to-Noise Ratio (PSNR) and perceptual similarity (LPIPS)~\cite{zhang2018perceptual}  as standard metrics for novel view synthesis. 
To evaluate the extrapolation capacities in regard of image fidelity, we evaluate Fr\'{e}chet Inception Distance~\cite{FID} (FID) and  Kernel Inception Distance~\cite{KID} (KID) for long-term settings.
To measure the video-consistency of the generated trajectory images, we compute Fr\'{e}chet Video Distance (FVD)~\cite{unterthiner2018FVD} scores.
Further, we follow ~\cite{Yu2023PhotoconsistentNVS} and report the symmetric epipolar distance (SED) to quantify faithfulness with respect to the provided camera trajectory, i.e., relative pose accuracy. 
Here, we compute the mean thresholded symmetric epipolar distance (mTSED) over the pixel thresholds $[1.0, 1.5, 2.0, 2.5, 3.0, 3.5, 4.0]$ and refer to the supplementary for detailed results. 

\begin{figure*}
\begin{center}
\includegraphics[width=\textwidth, trim={0.1cm 0 4.5cm 0cm},clip, page=3]{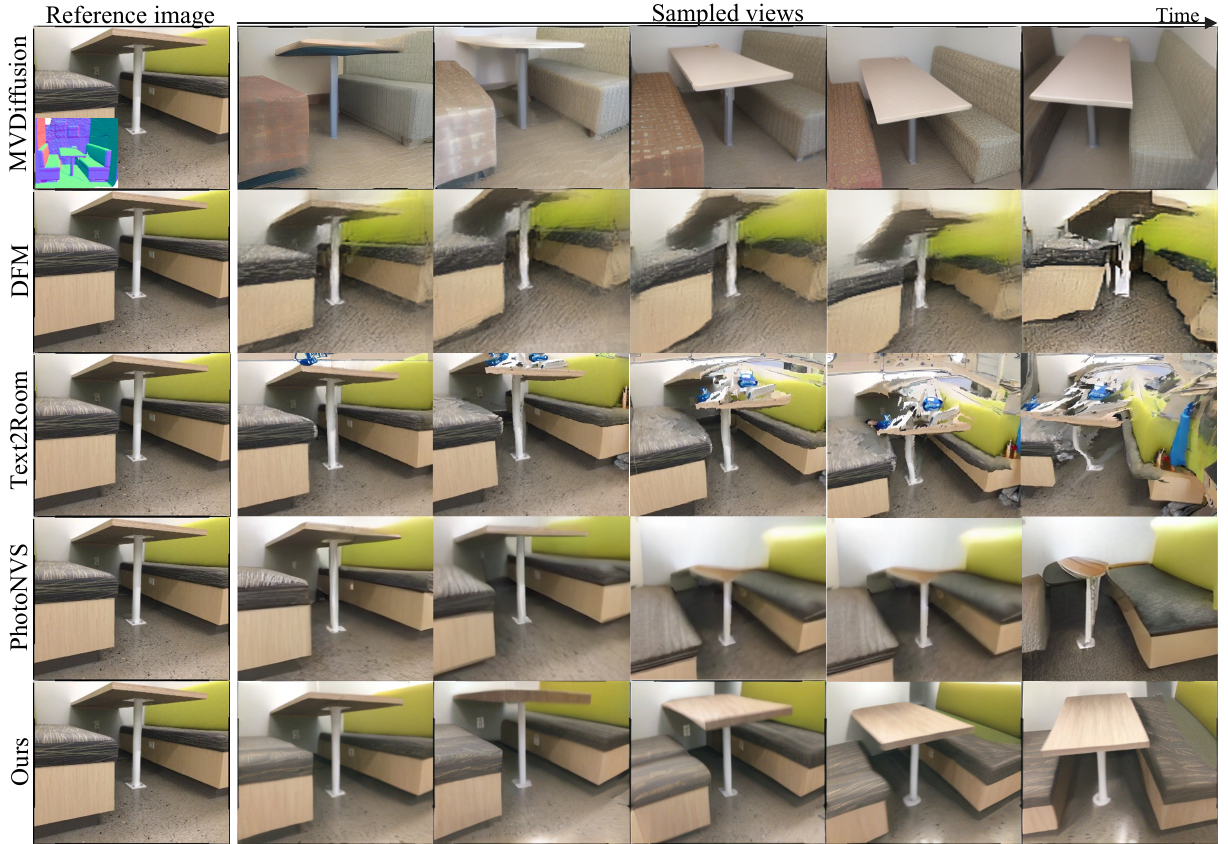}
\end{center}
\vspace{-1em}
\caption{
Generated views along ScanNet~\cite{dai2017scannet} test sequence (right) given a reference view (left). Our method simultaneously generates sequences of novel views that are both more realistic and more view-consistent than the baselines, DFM and PhotoNVS, which suffer from a considerable performance drop across large view point changes. Although MVDiffusion uses sensor depth input, the generated views are much less consistent with the reference image (e.g., colors of the cushions), compared to our generations, which do not rely on sensor depth.\looseness=-1
}
\label{fig:results_scannet}
\vspace{-5mm}
\end{figure*}
\vspace{-1em}
\paragraph{Baselines}
 We compare our approach against the state-of-the-art approaches for scene synthesis from a single reference image, including \DFM~\cite{tewari2023diffusion}, \PhotoNVS~\cite{Yu2023PhotoconsistentNVS}, \TR~\cite{hoellein2023text2room} and \MVD~\cite{tang2023MVDiffusion}. As \MVD is purely text-conditional, we incorporate the reference image during inference as follows. We use DDIM inversion to obtain the noise corresponding to the reference image and include it in the batch during sampling. Due to the global awareness of \MVD, information from the reference image can propagate to all generated views. Please see \SUPP for more details.
DFM trains a diffusion model to directly sample from the distribution of 3D scenes. 
Unlike our approach, \DFM cannot directly integrate 2D diffusion priors and does not generalize well to out-of-domain inputs as our experiments on ScanNet indicate.
PhotoNVS~\cite{Yu2023PhotoconsistentNVS} trains a pose-conditioned 2D diffusion model to iteratively predict the next frame for a given camera trajectory. \TR~\cite{hoellein2023text2room} uses an auto-regressive approach of predicting depth and leveraging a depth-conditional T2I model to generate new views that are used to update a textured mesh. 
In contrast, \Ours generates multiple frames from the input image in parallel, resulting in better long-term view synthesis and faster inference: To synthesize a $128\times 128$ frame, \PhotoNVS requires $\approx 45$s, \DFM $\approx 17$s while ours only takes $\approx 1$s .

\subsection{Consistent Novel View Synthesis}
\paragraph{Comparison against state of the art}
We quantitatively evaluate our approach on the task of consistent novel-view synthesis from a single reference on RealEstate10K~\cite{realestate46965} in \cref{tab:re10k} and ScanNet in \cref{tab:scannet}. 
Since \DFM does not support higher resolutions than 128px due to memory limitations, whereas the other methods run at a default resolution of 256px, we perform separate analyses at both resolutions. 

On RealEstate10K, we observe that our method achieves consistently better FID and KID scores on both short-term, as well as long-term evaluations: The short-term FID compared to DFM improves from $36.37$ to $25.30$ (at 128px), while the long-term FID improves by $33\%$ compared to \PhotoNVS. 
Moreover, our model outperforms all baselines in terms of FVD and achieves comparable results on LPIPS and mTSED with respect to DFM. 
We note that the PixelNeRF~\cite{yu2021pixelnerf} representation of DFM leads to highly consistent results, therefore good scores on pixel-level metrics like short-term PSNR, however, this comes at the cost of sharpness (as reflected in FID/KID). 

By leveraging strong image- and video-diffusion priors, our method achieves clear improvements over the baselines on ScanNet: 
As shown in~\cref{tab:scannet}, \Ours outperforms \MVD on short- and long-term metrics, indicating our model's ability to learn long-term correspondences even without relying on ground-truth geometry. 
In comparison to \DFM, \TR and \PhotoNVS, we observe strong photometric short- and long-term improvements over all baselines. \cref{fig:re10k_results,fig:results_scannet} show qualitative comparisons on RealEstate10K and ScanNet, respectively. It stands out that our method synthesises realistic and consistent novel views even across large viewpoint changes, where the quality of the baselines drops noticeable.

\subsection{Ablations}\label{sec:ablations}

\begin{table}
    \centering
    \resizebox{.99\columnwidth}{!}{
    \begin{tabular}{l|ccc|ccc}
          \multirow{2}{*}{Method} & \multicolumn{3}{c|}{Short-term} & \multicolumn{3}{c}{Long-term} \\
          & PSNR $\uparrow$ & LPIPS ↓ & FID ↓ & FID ↓  & FVD  ↓ & mTSED  $\uparrow$  \\
         \hline
         \Ours no prior & 14.29 & 0.493 & 63.56 & 85.30 & 236.8 & 0.587 \\
         \Ours no vid. & 14.68 & 0.552 & 37.05 & 38.43 & 214.9 & 0.728 \\
          \Ours no warp & 13.65 & 0.557 & 47.42 & 58.30 & 181.1 & 0.484 \\
         \Ours no pose & 15.53 & 0.417 & 27.84 & 32.25 & 120.26 & 0.624 \\
         \hline
        \Ours (Ours)& \textbf{15.65} & \textbf{0.393} & \textbf{25.90} & \textbf{30.15} &  \textbf{105.9} &  \textbf{0.855}
        
    \end{tabular}}
    \caption{Ablation of individual components of our pipeline on RealEstate10K~\cite{realestate46965} test sequences at 256px resolution.} 
    \label{tab:ablation2}
          \vspace{-4mm}
\end{table}

We show the contributions of individual components of our approach in \cref{tab:ablation2} and refer to the supplementary material for more qualitative comparisons. 
\vspace{-1em}
\paragraph{Importance of priors}
As described in \cref{sec:video_prior}, we initialize our model with weights obtained by training on large-scale image and video datasets. 
To study the importance of those priors for the task of consistent novel-view synthesis from a single image, we ablate them one by one: 
As shown in \cref{tab:ablation2}, training from scratch ("\Ours no prior") leads to strong degradation of image quality, as well as overall consistency.
Removing the video diffusion prior ("\Ours no vid.") has strong influence on the long-term consistency (mTSED decreases by $12.7\%$), as well as the video quality (FVD increases by more than $120\%$). 
We further ablate the monocular depth estimates on the reference image as condition to our model in "\Ours no warp" (\cref{tab:ablation2}). The drop in mTSED from $85.5\%$ to $48.4\%$ indicates that the model without reference warps is not able to closely adhere to the input trajectory. Besides the depth-warpings of the reference image, our method uses relative camera poses to synthesize images from the desired target poses.
When removing this modality ("\Ours no pose"), we notice effect on long-term generation becomes apparent, where there is minimal to no warp-guidance to inform about the desired camera poses, hence mTSED decreases from $0.85$ to $0.62$. 

\begin{figure}[ht]
\begin{center}
\includegraphics[width=.99\columnwidth, page=2, trim={0.3cm 2.2cm 7.8cm 0cm},clip]{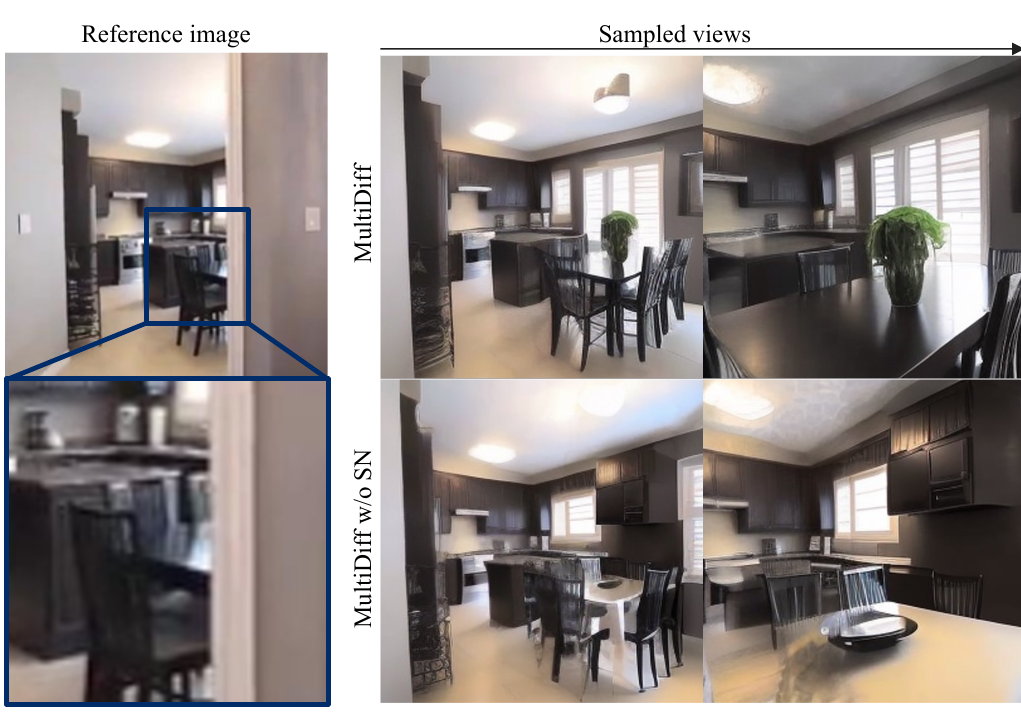}
\end{center}
\vspace{-2em}
\caption{
Without structured noise ("\Ours w/o SN"), the color of the dining table is not maintained \wrt the reference image.  
}
\label{fig:noise_warping}
\vspace{-3mm}
\end{figure}

\paragraph{Importance of structured noise}
As described in~\cref{sec:struct_noise}, we introduce structured noise by warping the initial noise consistently between target views according the depth estimates of the reference image. We measure the effect of noise warping in~\cref{tab:re10k} and~\cref{tab:scannet} ("\Ours w/o SN") on RealEstate10K and ScanNet trajectories. On both datasets, we observe that the structured noise leads to significantly more consistent and higher quality synthesis results. We show the effect of noise-warping in~\cref{fig:noise_warping}.

\subsection{Consistent Editing}

\begin{figure}
\begin{center}
\includegraphics[width=.99\columnwidth, page=1, trim={0.3cm 2.2cm 8.8cm 0cm},clip]{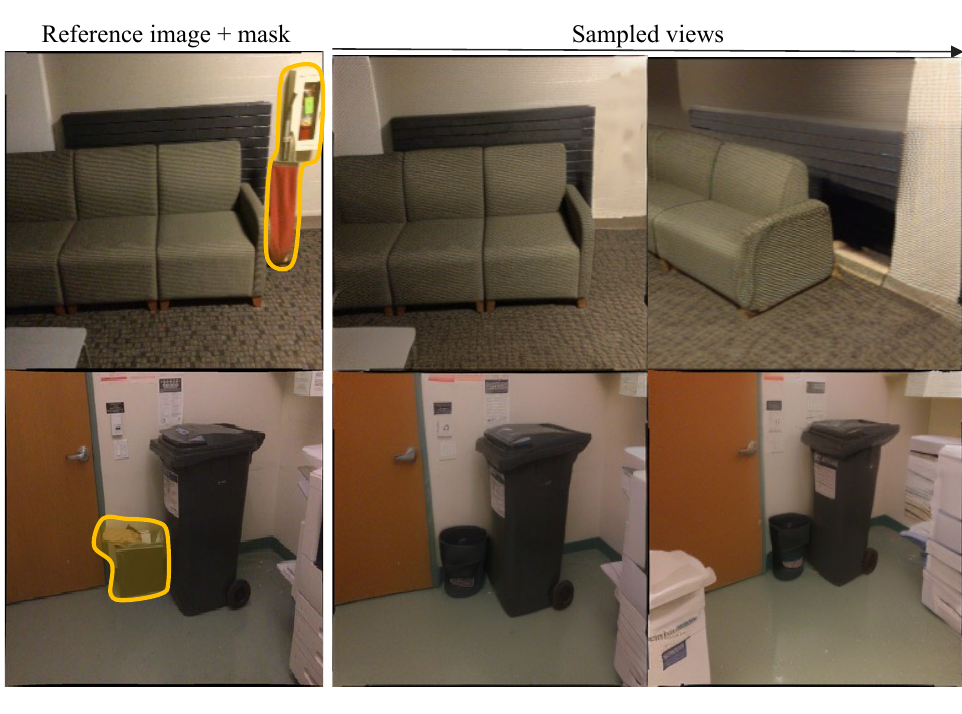}
\end{center}
\vspace{-2em}
\caption{
Consistent masking-based editing results on ScanNet test images.
}
\label{fig:editing}
\vspace{-3mm}
\end{figure}

In contrast to existing works such as \DFM or \PhotoNVS, our approach directly supports consistent editing without task-specific training.
During training, our model is tasked to synthesize consistent novel views even in absence of meaningful reference warps. 
By masking an area in a reference image that should not be warped, our model naturally performs consistent completion in those regions. 
We show examples on ScanNet test images in \cref{fig:editing} and refer to the supplementary material for more qualitative results.

\section{Conclusions}
\vspace{-2mm}
\label{sec:conclusions}
In this paper, we introduce \Ours, a novel approach for view extrapolation from a single input image. 
We identify video priors as a powerful proxy for this setting and demonstrate how they can be incorporated and adapted by converting temporal attention to \textit{correspondence attention}. With monocular depth cues, we facilitate learning improved correspondences by conditioning our model on reference views warped \wrt the target camera trajectory. 
Our experiments on RealEstate10k and ScanNet show significant improvements over relevant baselines, with particular gains on long-term sequence generation and overall inference speed.

\section*{Acknowledgements}
\vspace{-1mm}
Matthias Nie{\ss}ner was supported by the ERC Starting Grant Scan2CAD (804724). 
\clearpage
{
    \small
    \bibliographystyle{ieeenat_fullname}
    \bibliography{main}

\begin{thebibliography}{86}
\providecommand{\natexlab}[1]{#1}
\providecommand{\url}[1]{\texttt{#1}}
\expandafter\ifx\csname urlstyle\endcsname\relax
  \providecommand{\doi}[1]{doi: #1}\else
  \providecommand{\doi}{doi: \begingroup \urlstyle{rm}\Url}\fi

\bibitem[Anciukevicius et~al.(2022)Anciukevicius, Xu, Fisher, Henderson, Bilen, Mitra, and Guerrero]{anciukevicius2022renderdiffusion}
Titas Anciukevicius, Zexiang Xu, Matthew Fisher, Paul Henderson, Hakan Bilen, Niloy~J. Mitra, and Paul Guerrero.
\newblock {RenderDiffusion}: Image diffusion for {3D} reconstruction, inpainting and generation.
\newblock \emph{arXiv}, 2022.

\bibitem[Bae et~al.(2022)Bae, Budvytis, and Cipolla]{Bae2022}
Gwangbin Bae, Ignas Budvytis, and Roberto Cipolla.
\newblock Irondepth: Iterative refinement of single-view depth using surface normal and its uncertainty.
\newblock In \emph{British Machine Vision Conference (BMVC)}, 2022.

\bibitem[Bain et~al.(2021)Bain, Nagrani, Varol, and Zisserman]{Bain21}
Max Bain, Arsha Nagrani, G{\"u}l Varol, and Andrew Zisserman.
\newblock Frozen in time: A joint video and image encoder for end-to-end retrieval.
\newblock In \emph{IEEE International Conference on Computer Vision}, 2021.

\bibitem[Balaji et~al.(2022)Balaji, Nah, Huang, Vahdat, Song, Kreis, Aittala, Aila, Laine, Catanzaro, et~al.]{balaji2022ediffi}
Yogesh Balaji, Seungjun Nah, Xun Huang, Arash Vahdat, Jiaming Song, Karsten Kreis, Miika Aittala, Timo Aila, Samuli Laine, Bryan Catanzaro, et~al.
\newblock ediffi: Text-to-image diffusion models with an ensemble of expert denoisers.
\newblock \emph{arXiv preprint arXiv:2211.01324}, 2022.

\bibitem[Bautista et~al.(2022)Bautista, Guo, Abnar, Talbott, Toshev, Chen, Dinh, Zhai, Goh, Ulbricht, Dehghan, and Susskind]{bautista2022gaudi}
Miguel~{\'A}ngel Bautista, Pengsheng Guo, Samira Abnar, Walter Talbott, Alexander~T Toshev, Zhuoyuan Chen, Laurent Dinh, Shuangfei Zhai, Hanlin Goh, Daniel Ulbricht, Afshin Dehghan, and Joshua~M. Susskind.
\newblock {GAUDI}: A neural architect for immersive 3{D} scene generation.
\newblock \emph{arXiv preprint arXiv:2207.13751}, 2022.

\bibitem[Bhat et~al.(2023)Bhat, Birkl, Wofk, Wonka, and Müller]{ZoeDepth}
Shariq~Farooq Bhat, Reiner Birkl, Diana Wofk, Peter Wonka, and Matthias Müller.
\newblock Zoedepth: Zero-shot transfer by combining relative and metric depth, 2023.

\bibitem[Bińkowski et~al.(2018)Bińkowski, Sutherland, Arbel, and Gretton]{KID}
Mikołaj Bińkowski, Dougal~J. Sutherland, Michael Arbel, and Arthur Gretton.
\newblock Demystifying {MMD} {GAN}s.
\newblock In \emph{International Conference on Learning Representations}, 2018.

\bibitem[Chan et~al.(2023)Chan, Nagano, Chan, Bergman, Park, Levy, Aittala, Mello, Karras, and Wetzstein]{chan2023genvs}
Eric~R. Chan, Koki Nagano, Matthew~A. Chan, Alexander~W. Bergman, Jeong~Joon Park, Axel Levy, Miika Aittala, Shalini~De Mello, Tero Karras, and Gordon Wetzstein.
\newblock {GeNVS}: Generative novel view synthesis with {3D}-aware diffusion models.
\newblock In \emph{CVPR}, 2023.

\bibitem[Chen et~al.(2021)Chen, Xu, Zhao, Zhang, Xiang, Yu, and Su]{chen2021mvsnerf}
Anpei Chen, Zexiang Xu, Fuqiang Zhao, Xiaoshuai Zhang, Fanbo Xiang, Jingyi Yu, and Hao Su.
\newblock Mvsnerf: Fast generalizable radiance field reconstruction from multi-view stereo.
\newblock In \emph{CVPR}, 2021.

\bibitem[contributors()]{videocrafter}
VideoCrafter contributors.
\newblock Videocrafter.
\newblock \emph{Github}. Accessed October 15, 2023 [Online] \url{https://github.com/AILab-CVC/VideoCrafter}.

\bibitem[Dai et~al.(2017)Dai, Chang, Savva, Halber, Funkhouser, and Nie{\ss}ner]{dai2017scannet}
Angela Dai, Angel~X Chang, Manolis Savva, Maciej Halber, Thomas Funkhouser, and Matthias Nie{\ss}ner.
\newblock Scannet: Richly-annotated 3d reconstructions of indoor scenes.
\newblock In \emph{Proceedings of the IEEE conference on computer vision and pattern recognition}, pages 5828--5839, 2017.

\bibitem[Dai et~al.(2020)Dai, Zhang, Li, Liu, and Zeng]{dai2020pcrender}
Peng Dai, Yinda Zhang, Zhuwen Li, Shuaicheng Liu, and Bing Zeng.
\newblock Neural point cloud rendering via multi-plane projection.
\newblock In \emph{CVPR}, 2020.

\bibitem[Dhariwal and Nichol(2021)]{dhariwal2021diffusion}
Prafulla Dhariwal and Alexander Nichol.
\newblock Diffusion models beat gans on image synthesis.
\newblock \emph{Advances in Neural Information Processing Systems}, 34:\penalty0 8780--8794, 2021.

\bibitem[Dockhorn et~al.(2022)Dockhorn, Vahdat, and Kreis]{dockhorn2022langevin}
Tim Dockhorn, Arash Vahdat, and Karsten Kreis.
\newblock Score-based generative modeling with critically-damped langevin diffusion.
\newblock In \emph{ICLR}, 2022.

\bibitem[Du et~al.(2023)Du, Smith, Tewari, and Sitzmann]{du2023wide}
Yilun Du, Cameron Smith, Ayush Tewari, and Vincent Sitzmann.
\newblock Learning to render novel views from wide-baseline stereo pairs.
\newblock In \emph{CVPR}, 2023.

\bibitem[Esser et~al.(2023)Esser, Chiu, Atighehchian, Granskog, and Germanidis]{esser2023structure}
Patrick Esser, Johnathan Chiu, Parmida Atighehchian, Jonathan Granskog, and Anastasis Germanidis.
\newblock Structure and content-guided video synthesis with diffusion models.
\newblock In \emph{ICCV}, 2023.

\bibitem[Fridman et~al.(2023)Fridman, Abecasis, Kasten, and Dekel]{fridman2023SceneScape}
Rafail Fridman, Amit Abecasis, Yoni Kasten, and Tali Dekel.
\newblock Scenescape: Text-driven consistent scene generation.
\newblock In \emph{NeurIPS}, 2023.

\bibitem[Gu et~al.(2022)Gu, Chen, Bao, Wen, Zhang, Chen, Yuan, and Guo]{gu2022vector}
Shuyang Gu, Dong Chen, Jianmin Bao, Fang Wen, Bo Zhang, Dongdong Chen, Lu Yuan, and Baining Guo.
\newblock Vector quantized diffusion model for text-to-image synthesis.
\newblock In \emph{CVPR}, 2022.

\bibitem[Gu et~al.(2023)Gu, Wen, Song, and Gao]{gu2023seer}
Xianfan Gu, Chuan Wen, Jiaming Song, and Yang Gao.
\newblock Seer: Language instructed video prediction with latent diffusion models.
\newblock \emph{arXiv preprint arXiv:2303.14897}, 2023.

\bibitem[He et~al.(2023)He, Yang, Chen, Cun, Xia, Zhang, Wang, He, Chen, and Shan]{he2023scalecrafter}
Yingqing He, Shaoshu Yang, Haoxin Chen, Xiaodong Cun, Menghan Xia, Yong Zhang, Xintao Wang, Ran He, Qifeng Chen, and Ying Shan.
\newblock Scalecrafter: Tuning-free higher-resolution visual generation with diffusion models.
\newblock \emph{arXiv preprint arXiv:2310.07702}, 2023.

\bibitem[Henzler et~al.(2021)Henzler, Reizenstein, Labatut, Shapovalov, Ritschel, Vedaldi, and Novotny]{Henzler_2021_CVPR}
Philipp Henzler, Jeremy Reizenstein, Patrick Labatut, Roman Shapovalov, Tobias Ritschel, Andrea Vedaldi, and David Novotny.
\newblock Unsupervised learning of 3d object categories from videos in the wild.
\newblock In \emph{Proceedings of the IEEE/CVF Conference on Computer Vision and Pattern Recognition (CVPR)}, pages 4700--4709, 2021.

\bibitem[Heusel et~al.(2017)Heusel, Ramsauer, Unterthiner, Nessler, and Hochreiter]{FID}
Martin Heusel, Hubert Ramsauer, Thomas Unterthiner, Bernhard Nessler, and Sepp Hochreiter.
\newblock {GAN}s trained by a two time-scale update rule converge to a local nash equilibrium.
\newblock \emph{Advances in neural information processing systems}, 30, 2017.

\bibitem[Ho et~al.(2020)Ho, Jain, and Abbeel]{ho2020denoising}
Jonathan Ho, Ajay Jain, and Pieter Abbeel.
\newblock Denoising diffusion probabilistic models.
\newblock \emph{NeurIPS}, 2020.

\bibitem[Ho et~al.(2022{\natexlab{a}})Ho, Chan, Saharia, Whang, Gao, Gritsenko, Kingma, Poole, Norouzi, Fleet, et~al.]{ho2022imagenvideo}
Jonathan Ho, William Chan, Chitwan Saharia, Jay Whang, Ruiqi Gao, Alexey Gritsenko, Diederik~P Kingma, Ben Poole, Mohammad Norouzi, David~J Fleet, et~al.
\newblock Imagen video: High definition video generation with diffusion models.
\newblock \emph{arXiv preprint arXiv:2210.02303}, 2022{\natexlab{a}}.

\bibitem[Ho et~al.(2022{\natexlab{b}})Ho, Saharia, Chan, Fleet, Norouzi, and Salimans]{ho2022cascaded}
Jonathan Ho, Chitwan Saharia, William Chan, David~J Fleet, Mohammad Norouzi, and Tim Salimans.
\newblock Cascaded diffusion models for high fidelity image generation.
\newblock 23:\penalty0 47--1, 2022{\natexlab{b}}.

\bibitem[Ho et~al.(2022{\natexlab{c}})Ho, Salimans, Gritsenko, Chan, Norouzi, and Fleet]{ho2022video}
Jonathan Ho, Tim Salimans, Alexey Gritsenko, William Chan, Mohammad Norouzi, and David~J Fleet.
\newblock Video diffusion models.
\newblock \emph{ICLR}, 2022{\natexlab{c}}.

\bibitem[H\"ollein et~al.(2023)H\"ollein, Cao, Owens, Johnson, and Nie{\ss}ner]{hoellein2023text2room}
Lukas H\"ollein, Ang Cao, Andrew Owens, Justin Johnson, and Matthias Nie{\ss}ner.
\newblock Text2room: Extracting textured 3d meshes from 2d text-to-image models.
\newblock In \emph{ICCV}, 2023.

\bibitem[Kim et~al.(2023)Kim, Brown, Yin, Kreis, Schwarz, Li, Rombach, Torralba, and Fidler]{kim2023nfldm}
Seung~Wook Kim, Bradley Brown, Kangxue Yin, Karsten Kreis, Katja Schwarz, Daiqing Li, Robin Rombach, Antonio Torralba, and Sanja Fidler.
\newblock Neuralfield-ldm: Scene generation with hierarchical latent diffusion models.
\newblock In \emph{CVPR}, 2023.

\bibitem[Kingma and Ba(2015)]{kingma2015adam}
Diederik~P. Kingma and Jimmy Ba.
\newblock Adam: {A} method for stochastic optimization.
\newblock In \emph{ICLR}, 2015.

\bibitem[Koh et~al.(2021)Koh, Lee, Yang, Baldridge, and Anderson]{koh2021pathdreamer}
Jing~Yu Koh, Honglak Lee, Yinfei Yang, Jason Baldridge, and Peter Anderson.
\newblock Pathdreamer: {A} world model for indoor navigation.
\newblock In \emph{ICCV}, 2021.

\bibitem[Koh et~al.(2023)Koh, Agrawal, Batra, Tucker, Waters, Lee, Yang, Baldridge, and Anderson]{koh2022simple}
Jing~Yu Koh, Harsh Agrawal, Dhruv Batra, Richard Tucker, Austin Waters, Honglak Lee, Yinfei Yang, Jason Baldridge, and Peter Anderson.
\newblock Simple and effective synthesis of indoor {3D} scenes.
\newblock \emph{AAAI}, 2023.

\bibitem[Kulh{\'{a}}nek et~al.(2022)Kulh{\'{a}}nek, Derner, Sattler, and Babuska]{kulhanek2022viewformer}
Jon{\'{a}}s Kulh{\'{a}}nek, Erik Derner, Torsten Sattler, and Robert Babuska.
\newblock Viewformer: Nerf-free neural rendering from few images using transformers.
\newblock In \emph{ECCV}, 2022.

\bibitem[Li et~al.(2022)Li, Wang, Snavely, and Kanazawa]{li2022infinite}
Zhengqi Li, Qianqian Wang, Noah Snavely, and Angjoo Kanazawa.
\newblock Infinitenature-zero: Learning perpetual view generation of natural scenes from single images.
\newblock In \emph{ECCV}, 2022.

\bibitem[Liu et~al.(2021)Liu, Tucker, Jampani, Makadia, Snavely, and Kanazawa]{liu2021infinite}
Andrew Liu, Richard Tucker, Varun Jampani, Ameesh Makadia, Noah Snavely, and Angjoo Kanazawa.
\newblock Infinite nature: Perpetual view generation of natural scenes from a single image.
\newblock In \emph{ICCV}, 2021.

\bibitem[Liu et~al.(2023)Liu, Wu, Hoorick, Tokmakov, Zakharov, and Vondrick]{liu2023zero1to3}
Ruoshi Liu, Rundi Wu, Basile~Van Hoorick, Pavel Tokmakov, Sergey Zakharov, and Carl Vondrick.
\newblock Zero-1-to-3: Zero-shot one image to 3d object.
\newblock In \emph{ICCV}, 2023.

\bibitem[Lombardi et~al.(2019)Lombardi, Simon, Saragih, Schwartz, Lehrmann, and Sheikh]{lombardi2019neural}
Stephen Lombardi, Tomas Simon, Jason Saragih, Gabriel Schwartz, Andreas Lehrmann, and Yaser Sheikh.
\newblock Neural volumes: Learning dynamic renderable volumes from images.
\newblock \emph{arXiv preprint arXiv:1906.07751}, 2019.

\bibitem[Luo et~al.(2023)Luo, Chen, Zhang, Huang, Wang, Shen, Zhao, Zhou, and Tan]{luo2023videofusion}
Zhengxiong Luo, Dayou Chen, Yingya Zhang, Yan Huang, Liang Wang, Yujun Shen, Deli Zhao, Jingren Zhou, and Tieniu Tan.
\newblock Videofusion: Decomposed diffusion models for high-quality video generation.
\newblock In \emph{CVPR}, 2023.

\bibitem[Meshry et~al.(2019)Meshry, Goldman, Khamis, Hoppe, Pandey, Snavely, and Martin{-}Brualla]{meshry2019wild}
Moustafa Meshry, Dan~B. Goldman, Sameh Khamis, Hugues Hoppe, Rohit Pandey, Noah Snavely, and Ricardo Martin{-}Brualla.
\newblock Neural rerendering in the wild.
\newblock In \emph{CVPR}, 2019.

\bibitem[Mildenhall et~al.(2020)Mildenhall, Srinivasan, Tancik, Barron, Ramamoorthi, and Ng]{mildenhall2020nerf}
Ben Mildenhall, Pratul~P Srinivasan, Matthew Tancik, Jonathan~T Barron, Ravi Ramamoorthi, and Ren Ng.
\newblock Nerf: Representing scenes as neural radiance fields for view synthesis.
\newblock In \emph{European conference on computer vision}, pages 405--421. Springer, 2020.

\bibitem[M{\"{u}}ller et~al.(2022)M{\"{u}}ller, Simonelli, Porzi, Bulo, Nie{\ss}ner, and Kontschieder]{mueller2022autorf}
Norman M{\"{u}}ller, Andrea Simonelli, Lorenzo Porzi, Samuel~Rota Bulo, Matthias Nie{\ss}ner, and Peter Kontschieder.
\newblock Autorf: Learning 3d object radiance fields from single view observations.
\newblock In \emph{Proceedings of the IEEE/CVF Conference on Computer Vision and Pattern Recognition (CVPR)}, 2022.

\bibitem[M{\"u}ller et~al.(2023)M{\"u}ller, Siddiqui, Porzi, Bulo, Kontschieder, and Nie{\ss}ner]{muller2023diffrf}
Norman M{\"u}ller, Yawar Siddiqui, Lorenzo Porzi, Samuel~Rota Bulo, Peter Kontschieder, and Matthias Nie{\ss}ner.
\newblock Diffrf: Rendering-guided 3d radiance field diffusion.
\newblock In \emph{Proceedings of the IEEE/CVF Conference on Computer Vision and Pattern Recognition}, pages 4328--4338, 2023.

\bibitem[Nichol and Dhariwal(2021)]{nichol2021improved}
Alexander~Quinn Nichol and Prafulla Dhariwal.
\newblock Improved denoising diffusion probabilistic models.
\newblock In \emph{International Conference on Machine Learning}, pages 8162--8171. PMLR, 2021.

\bibitem[Nichol et~al.(2022)Nichol, Dhariwal, Ramesh, Shyam, Mishkin, Mcgrew, Sutskever, and Chen]{nichol2022glide}
Alexander~Quinn Nichol, Prafulla Dhariwal, Aditya Ramesh, Pranav Shyam, Pamela Mishkin, Bob Mcgrew, Ilya Sutskever, and Mark Chen.
\newblock Glide: Towards photorealistic image generation and editing with text-guided diffusion models.
\newblock 2022.

\bibitem[Niemeyer et~al.(2020)Niemeyer, Mescheder, Oechsle, and Geiger]{niemeyer2020dvr}
Michael Niemeyer, Lars~M. Mescheder, Michael Oechsle, and Andreas Geiger.
\newblock Differentiable volumetric rendering: Learning implicit 3d representations without 3d supervision.
\newblock In \emph{CVPR}, 2020.

\bibitem[Niemeyer et~al.(2022)Niemeyer, Barron, Mildenhall, Sajjadi, Geiger, and Radwan]{niemeyer2022}
Michael Niemeyer, Jonathan~T. Barron, Ben Mildenhall, Mehdi S.~M. Sajjadi, Andreas Geiger, and Noha Radwan.
\newblock Regnerf: Regularizing neural radiance fields for view synthesis from sparse inputs.
\newblock In \emph{CVPR}, 2022.

\bibitem[Pavllo et~al.(2023)Pavllo, Tan, Rakotosaona, and Tombari]{pavllo2023shape}
Dario Pavllo, David~Joseph Tan, Marie{-}Julie Rakotosaona, and Federico Tombari.
\newblock Shape, pose, and appearance from a single image via bootstrapped radiance field inversion.
\newblock In \emph{CVPR}, 2023.

\bibitem[Podell et~al.(2023)Podell, English, Lacey, Blattmann, Dockhorn, M{\"{u}}ller, Penna, and Rombach]{podell2023sdxl}
Dustin Podell, Zion English, Kyle Lacey, Andreas Blattmann, Tim Dockhorn, Jonas M{\"{u}}ller, Joe Penna, and Robin Rombach.
\newblock {SDXL:} improving latent diffusion models for high-resolution image synthesis.
\newblock \emph{CoRR}, arxiv preprint arxiv:2307.01952, 2023.

\bibitem[Qiu et~al.(2023)Qiu, Xia, Zhang, He, Wang, Shan, and Liu]{qiu2023freenoise}
Haonan Qiu, Menghan Xia, Yong Zhang, Yingqing He, Xintao Wang, Ying Shan, and Ziwei Liu.
\newblock Freenoise: Tuning-free longer video diffusion via noise rescheduling, 2023.

\bibitem[Ramesh et~al.(2022)Ramesh, Dhariwal, Nichol, Chu, and Chen]{ramesh2022hierarchical}
Aditya Ramesh, Prafulla Dhariwal, Alex Nichol, Casey Chu, and Mark Chen.
\newblock Hierarchical text-conditional image generation with clip latents.
\newblock \emph{arXiv preprint arXiv:2204.06125}, 2022.

\bibitem[Ren and Wang(2022)]{ren2022look}
Xuanchi Ren and Xiaolong Wang.
\newblock Look outside the room: Synthesizing a consistent long-term {3D} scene video from a single image.
\newblock In \emph{CVPR}, 2022.

\bibitem[Rockwell et~al.(2021)Rockwell, Fouhey, and Johnson]{rockwell2021pixelsynth}
Chris Rockwell, David~F Fouhey, and Justin Johnson.
\newblock Pixelsynth: Generating a {3D}-consistent experience from a single image.
\newblock In \emph{ICCV}, 2021.

\bibitem[Roessle et~al.(2022)Roessle, Barron, Mildenhall, Srinivasan, and Nie{\ss}ner]{roessle2022depthpriorsnerf}
Barbara Roessle, Jonathan~T. Barron, Ben Mildenhall, Pratul~P. Srinivasan, and Matthias Nie{\ss}ner.
\newblock Dense depth priors for neural radiance fields from sparse input views.
\newblock In \emph{Proceedings of the IEEE/CVF Conference on Computer Vision and Pattern Recognition (CVPR)}, 2022.

\bibitem[Rombach et~al.(2021{\natexlab{a}})Rombach, Blattmann, Lorenz, Esser, and Ommer]{rombach2021highresolution}
Robin Rombach, Andreas Blattmann, Dominik Lorenz, Patrick Esser, and Björn Ommer.
\newblock High-resolution image synthesis with latent diffusion models, 2021{\natexlab{a}}.

\bibitem[Rombach et~al.(2021{\natexlab{b}})Rombach, Esser, and Ommer]{rombach2021geometry}
Robin Rombach, Patrick Esser, and Bj{\"o}rn Ommer.
\newblock Geometry-free view synthesis: Transformers and no {3D} priors.
\newblock In \emph{ICCV}, 2021{\natexlab{b}}.

\bibitem[Ross et~al.(2011)Ross, Gordon, and Bagnell]{pmlr-v15-ross11a}
Stephane Ross, Geoffrey Gordon, and Drew Bagnell.
\newblock A reduction of imitation learning and structured prediction to no-regret online learning.
\newblock In \emph{Int. Conf. Art. Intell. Stat.} PMLR, 2011.

\bibitem[Saharia et~al.(2022)Saharia, Chan, Saxena, Li, Whang, Denton, Ghasemipour, Gontijo~Lopes, Karagol~Ayan, Salimans, et~al.]{saharia2022photorealistic}
Chitwan Saharia, William Chan, Saurabh Saxena, Lala Li, Jay Whang, Emily~L Denton, Kamyar Ghasemipour, Raphael Gontijo~Lopes, Burcu Karagol~Ayan, Tim Salimans, et~al.
\newblock Photorealistic text-to-image diffusion models with deep language understanding.
\newblock In \emph{NeurIPS}, 2022.

\bibitem[Sajjadi et~al.(2022)Sajjadi, Meyer, Pot, Bergmann, Greff, Radwan, Vora, Lucic, Duckworth, Dosovitskiy, Uszkoreit, Funkhouser, and Tagliasacchi]{sajjadi2022scene}
Mehdi S.~M. Sajjadi, Henning Meyer, Etienne Pot, Urs Bergmann, Klaus Greff, Noha Radwan, Suhani Vora, Mario Lucic, Daniel Duckworth, Alexey Dosovitskiy, Jakob Uszkoreit, Thomas Funkhouser, and Andrea Tagliasacchi.
\newblock Scene representation transformer: Geometry-free novel view synthesis through set-latent scene representations.
\newblock In \emph{CVPR}, 2022.

\bibitem[Sargent et~al.(2023)Sargent, Li, Shah, Herrmann, Yu, Zhang, Chan, Lagun, Fei-Fei, Sun, and Wu]{sargent2023zeronvs}
Kyle Sargent, Zizhang Li, Tanmay Shah, Charles Herrmann, Hong-Xing Yu, Yunzhi Zhang, Eric~Ryan Chan, Dmitry Lagun, Li Fei-Fei, Deqing Sun, and Jiajun Wu.
\newblock {ZeroNVS}: Zero-shot 360-degree view synthesis from a single real image.
\newblock \emph{arXiv preprint arXiv:2310.17994}, 2023.

\bibitem[Singer et~al.(2023)Singer, Polyak, Hayes, Yin, An, Zhang, Hu, Yang, Ashual, Gafni, et~al.]{singer2022make}
Uriel Singer, Adam Polyak, Thomas Hayes, Xi Yin, Jie An, Songyang Zhang, Qiyuan Hu, Harry Yang, Oron Ashual, Oran Gafni, et~al.
\newblock Make-a-video: Text-to-video generation without text-video data.
\newblock In \emph{ICLR}, 2023.

\bibitem[Sitzmann et~al.(2019{\natexlab{a}})Sitzmann, Thies, Heide, Nie{\ss}ner, Wetzstein, and Zollh{\"{o}}fer]{sitzmann2019deepvoxels}
Vincent Sitzmann, Justus Thies, Felix Heide, Matthias Nie{\ss}ner, Gordon Wetzstein, and Michael Zollh{\"{o}}fer.
\newblock Deepvoxels: Learning persistent 3d feature embeddings.
\newblock In \emph{CVPR}, 2019{\natexlab{a}}.

\bibitem[Sitzmann et~al.(2019{\natexlab{b}})Sitzmann, Zollh{\"o}fer, and Wetzstein]{sitzmann2019scene}
Vincent Sitzmann, Michael Zollh{\"o}fer, and Gordon Wetzstein.
\newblock Scene representation networks: Continuous 3d-structure-aware neural scene representations.
\newblock \emph{arXiv preprint arXiv:1906.01618}, 2019{\natexlab{b}}.

\bibitem[Sohl-Dickstein et~al.(2015)Sohl-Dickstein, Weiss, Maheswaranathan, and Ganguli]{sohl2015deep}
Jascha Sohl-Dickstein, Eric Weiss, Niru Maheswaranathan, and Surya Ganguli.
\newblock Deep unsupervised learning using nonequilibrium thermodynamics.
\newblock pages 2256--2265. PMLR, 2015.

\bibitem[Sohl{-}Dickstein et~al.(2015)Sohl{-}Dickstein, Weiss, Maheswaranathan, and Ganguli]{sohldickstein2015deep}
Jascha Sohl{-}Dickstein, Eric~A. Weiss, Niru Maheswaranathan, and Surya Ganguli.
\newblock Deep unsupervised learning using nonequilibrium thermodynamics.
\newblock In \emph{ICML}, 2015.

\bibitem[Song et~al.(2020)Song, Meng, and Ermon]{song2020denoising}
Jiaming Song, Chenlin Meng, and Stefano Ermon.
\newblock Denoising diffusion implicit models.
\newblock \emph{arXiv preprint arXiv:2010.02502}, 2020.

\bibitem[Song and Ermon(2019)]{genmodel}
Yang Song and Stefano Ermon.
\newblock Generative modeling by estimating gradients of the data distribution.
\newblock \emph{CoRR}, abs/1907.05600, 2019.

\bibitem[Song and Ermon(2020)]{song2020improved}
Yang Song and Stefano Ermon.
\newblock Improved techniques for training score-based generative models.
\newblock \emph{Advances in neural information processing systems}, 33:\penalty0 12438--12448, 2020.

\bibitem[Song et~al.(2021)Song, Sohl-Dickstein, Kingma, Kumar, Ermon, and Poole]{song2020score}
Yang Song, Jascha Sohl-Dickstein, Diederik~P Kingma, Abhishek Kumar, Stefano Ermon, and Ben Poole.
\newblock Score-based generative modeling through stochastic differential equations.
\newblock \emph{arXiv preprint arXiv:2011.13456}, 2021.

\bibitem[Tang et~al.(2023)Tang, Zhang, Chen, Wang, and Yasutaka]{tang2023MVDiffusion}
Shitao Tang, Fuayng Zhang, Jiacheng Chen, Peng Wang, and Furukawa Yasutaka.
\newblock Mvdiffusion: Enabling holistic multi-view image generation with correspondence-aware diffusion.
\newblock \emph{arXiv preprint 2307.01097}, 2023.

\bibitem[Tewari et~al.(2023)Tewari, Yin, Cazenavette, Rezchikov, Tenenbaum, Durand, Freeman, and Sitzmann]{tewari2023diffusion}
Ayush Tewari, Tianwei Yin, George Cazenavette, Semon Rezchikov, Joshua~B. Tenenbaum, Frédo Durand, William~T. Freeman, and Vincent Sitzmann.
\newblock Diffusion with forward models: Solving stochastic inverse problems without direct supervision.
\newblock \emph{NeurIPS}, 2023.

\bibitem[Thies et~al.(2019)Thies, Zollh{\"{o}}fer, and Nie{\ss}ner]{thies2019deferred}
Justus Thies, Michael Zollh{\"{o}}fer, and Matthias Nie{\ss}ner.
\newblock Deferred neural rendering: image synthesis using neural textures.
\newblock \emph{ACM TOG}, 2019.

\bibitem[Trevithick and Yang(2021)]{trevithick2020GRF}
Alex Trevithick and Bo Yang.
\newblock Grf: Learning a general radiance field for 3d scene representation and rendering.
\newblock In \emph{ICCV}, 2021.

\bibitem[Tseng et~al.(2023)Tseng, Li, Kim, Alsisan, Huang, and Kopf]{tseng2023consistent}
Hung-Yu Tseng, Qinbo Li, Changil Kim, Suhib Alsisan, Jia-Bin Huang, and Johannes Kopf.
\newblock Consistent view synthesis with pose-guided diffusion models.
\newblock In \emph{CVPR}, 2023.

\bibitem[Unterthiner et~al.(2018)Unterthiner, van Steenkiste, Kurach, Marinier, Michalski, and Gelly]{unterthiner2018FVD}
Thomas Unterthiner, Sjoerd van Steenkiste, Karol Kurach, Raphael Marinier, Marcin Michalski, and Sylvain Gelly.
\newblock Towards accurate generative models of video: A new metric \& challenges.
\newblock \emph{arXiv preprint arXiv:1812.01717}, 2018.

\bibitem[Vahdat et~al.(2021)Vahdat, Kreis, and Kautz]{vahdat2021latent}
Arash Vahdat, Karsten Kreis, and Jan Kautz.
\newblock Score-based generative modeling in latent space.
\newblock In \emph{NeurIPS}, 2021.

\bibitem[Wang et~al.(2021)Wang, Wang, Genova, Srinivasan, Zhou, Barron, Martin-Brualla, Snavely, and Funkhouser]{wang2021ibrnet}
Qianqian Wang, Zhicheng Wang, Kyle Genova, Pratul Srinivasan, Howard Zhou, Jonathan~T. Barron, Ricardo Martin-Brualla, Noah Snavely, and Thomas Funkhouser.
\newblock Ibrnet: Learning multi-view image-based rendering.
\newblock In \emph{CVPR}, 2021.

\bibitem[Wang et~al.(2023)Wang, Yuan, Zhang, Chen, Wang, Zhang, Shen, Zhao, and Zhou]{wang2023videocomposer}
Xiang Wang, Hangjie Yuan, Shiwei Zhang, Dayou Chen, Jiuniu Wang, Yingya Zhang, Yujun Shen, Deli Zhao, and Jingren Zhou.
\newblock Videocomposer: Compositional video synthesis with motion controllability.
\newblock \emph{arXiv preprint arXiv:2306.02018}, 2023.

\bibitem[Watson et~al.(2023)Watson, Chan, Martin{-}Brualla, Ho, Tagliasacchi, and Norouzi]{watson20233dim}
Daniel Watson, William Chan, Ricardo Martin{-}Brualla, Jonathan Ho, Andrea Tagliasacchi, and Mohammad Norouzi.
\newblock Novel view synthesis with diffusion models.
\newblock In \emph{ICLR}, 2023.

\bibitem[Wiles et~al.(2020)Wiles, Gkioxari, Szeliski, and Johnson]{wiles2020synsin}
Olivia Wiles, Georgia Gkioxari, Richard Szeliski, and Justin Johnson.
\newblock {SynSin}: {E}nd-to-end view synthesis from a single image.
\newblock In \emph{CVPR}, 2020.

\bibitem[Wu et~al.(2023)Wu, Mildenhall, Henzler, Park, Gao, Watson, Srinivasan, Verbin, Barron, Poole, and Holynski]{wu2023reconfusion}
Rundi Wu, Ben Mildenhall, Philipp Henzler, Keunhong Park, Ruiqi Gao, Daniel Watson, Pratul~P. Srinivasan, Dor Verbin, Jonathan~T. Barron, Ben Poole, and Aleksander Holynski.
\newblock Reconfusion: 3d reconstruction with diffusion priors.
\newblock \emph{arXiv}, 2023.

\bibitem[Xiang et~al.(2023)Xiang, Yang, Huang, and Tong]{xiang2023ivid}
Jianfeng Xiang, Jiaolong Yang, Binbin Huang, and Xin Tong.
\newblock 3d-aware image generation using 2d diffusion models.
\newblock In \emph{ICCV}, 2023.

\bibitem[Yu et~al.(2021)Yu, Ye, Tancik, and Kanazawa]{yu2021pixelnerf}
Alex Yu, Vickie Ye, Matthew Tancik, and Angjoo Kanazawa.
\newblock {pixelNeRF}: Neural radiance fields from one or few images.
\newblock In \emph{CVPR}, pages 4578--4587, 2021.

\bibitem[Yu et~al.(2023)Yu, Forghani, Derpanis, and Brubaker]{Yu2023PhotoconsistentNVS}
Jason~J. Yu, Fereshteh Forghani, Konstantinos~G. Derpanis, and Marcus~A. Brubaker.
\newblock Long-term photometric consistent novel view synthesis with diffusion models.
\newblock In \emph{{Proceedings of the International Conference on Computer Vision ({ICCV})}}, 2023.

\bibitem[Zhang et~al.(2023)Zhang, Rao, and Agrawala]{zhang2023controlnet}
Lvmin Zhang, Anyi Rao, and Maneesh Agrawala.
\newblock Adding conditional control to text-to-image diffusion models, 2023.

\bibitem[Zhang et~al.(2018)Zhang, Isola, Efros, Shechtman, and Wang]{zhang2018perceptual}
Richard Zhang, Phillip Isola, Alexei~A Efros, Eli Shechtman, and Oliver Wang.
\newblock The unreasonable effectiveness of deep features as a perceptual metric.
\newblock In \emph{CVPR}, 2018.

\bibitem[Zhou et~al.(2018)Zhou, Tucker, Flynn, Fyffe, and Snavely]{realestate46965}
Tinghui Zhou, Richard Tucker, John Flynn, Graham Fyffe, and Noah Snavely.
\newblock Stereo magnification: Learning view synthesis using multiplane images.
\newblock In \emph{ACM TOG}, 2018.

\bibitem[Zhou and Tulsiani(2023)]{zhou2023sparsefusion}
Zhizhuo Zhou and Shubham Tulsiani.
\newblock Sparsefusion: Distilling view-conditioned diffusion for 3d reconstruction.
\newblock In \emph{CVPR}, 2023.

\end{thebibliography}
}

\newpage
\appendix

\newpage
\section*{\Large\textbf{Appendix}}

In this appendix, we provide additional qualitative and quantitative results and discuss training and evaluation details.

\section{Additional qualitative results}

\begin{figure*}
\begin{center}
\includegraphics[width=0.85\textwidth, trim={0cm 5.5cm 0.3cm 0cm},clip, page=2]{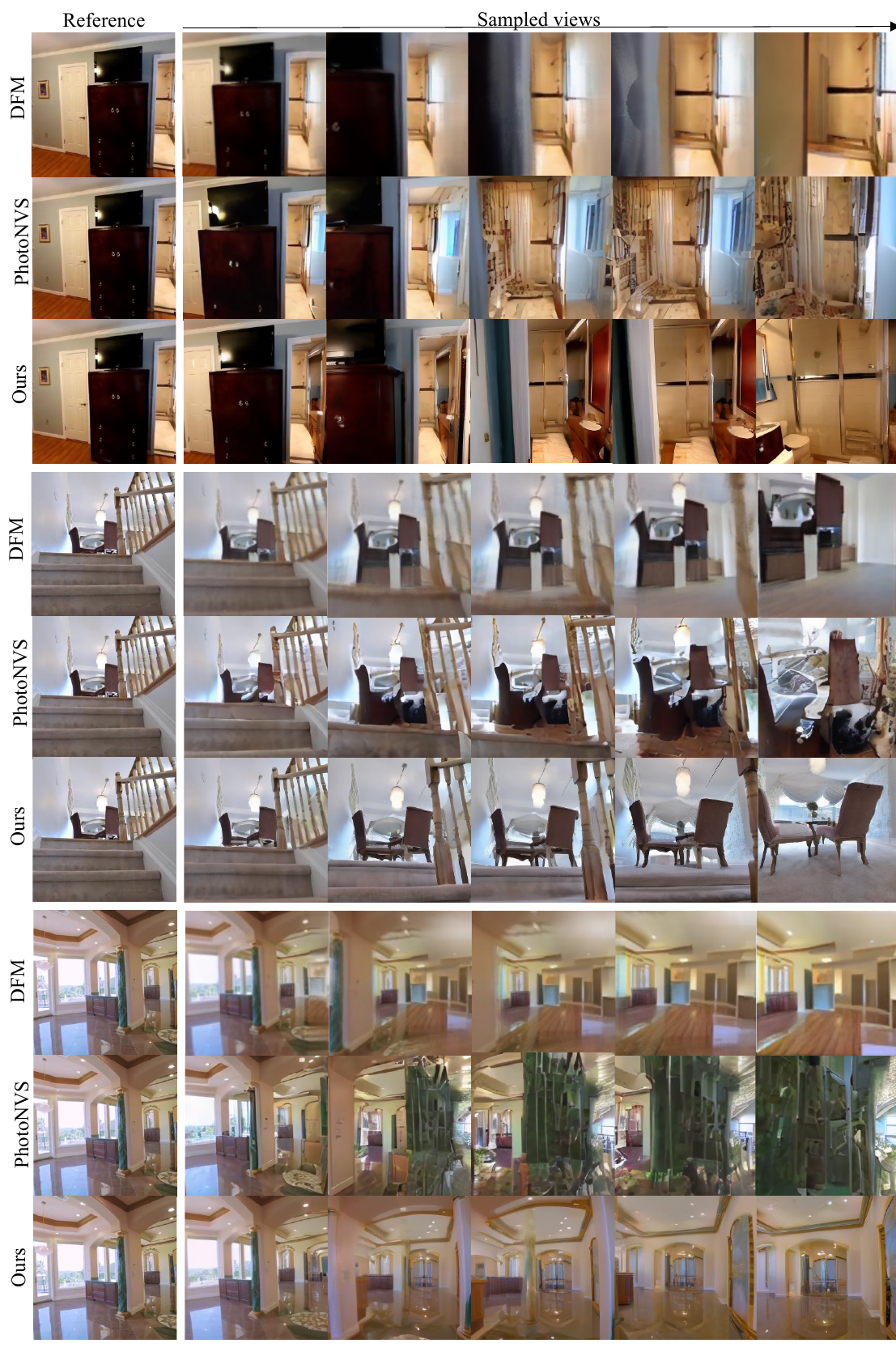}

\end{center}
\vspace{-1.5em}
\caption{
Additional qualitative comparison results on RealEstate10K~\cite{realestate46965}.
}
\label{fig:add_re10k_comp}

\end{figure*}

\begin{figure*}
\begin{center}
\includegraphics[width=0.88\textwidth, trim={0cm 8cm 0.0cm 0cm},clip, page=1]{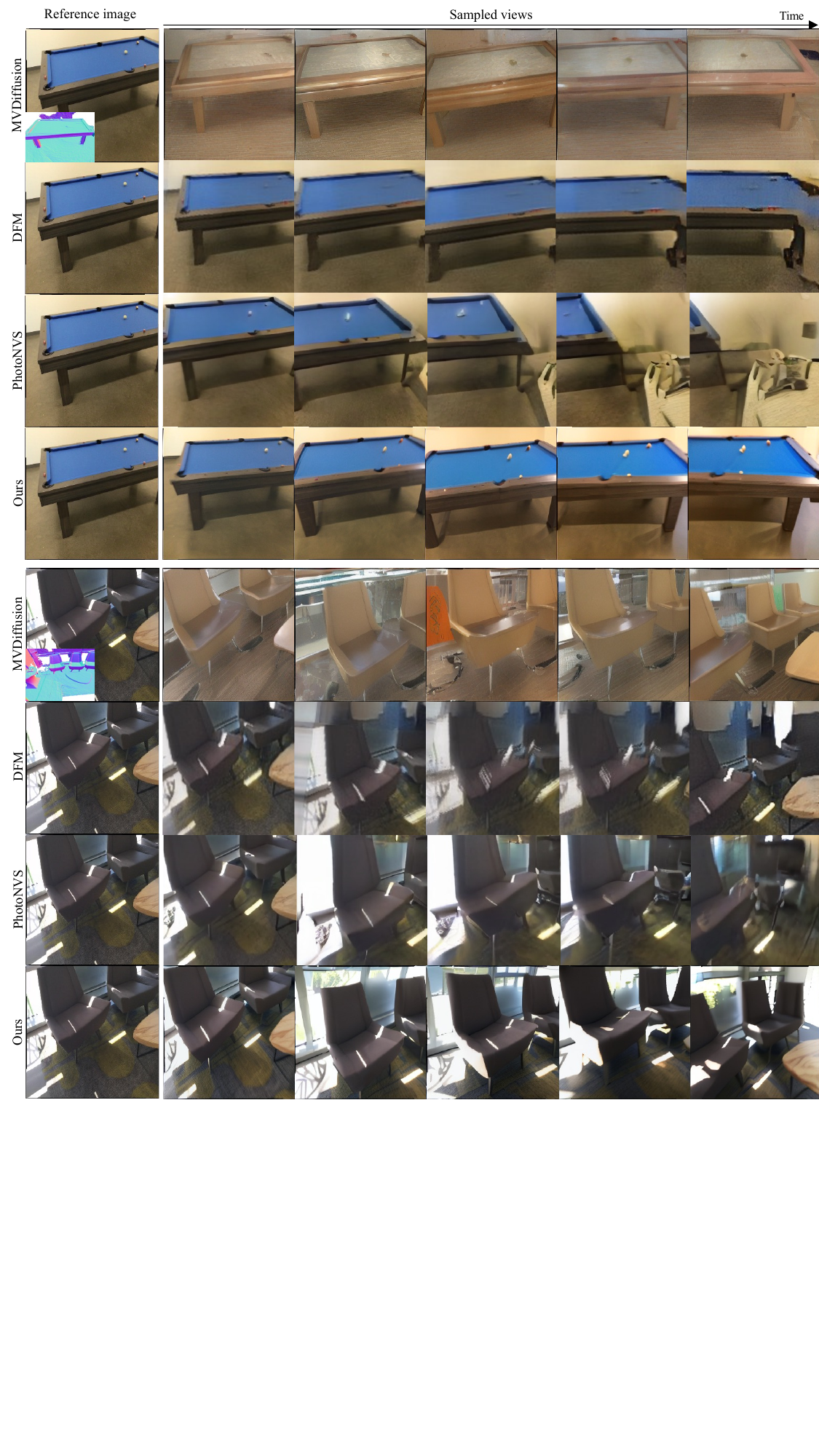}

\end{center}
\vspace{-1.5em}
\caption{
Additional qualitative comparison results on ScanNet~\cite{dai2017scannet}. Note that \MVD requires the scene mesh for inference.
}
\label{fig:add_scannet_comp}

\end{figure*}

We provide additional qualitative comparisons with the baselines on RealEstate10K~\cite{realestate46965} in \cref{fig:add_re10k_comp} as well as on ScanNet~\cite{dai2017scannet}  in \cref{fig:add_scannet_comp}. While \PhotoNVS~\cite{Yu2023PhotoconsistentNVS} accumulates errors over the autoregressive sampling process, our model synthesizes realistic images for all target poses jointly.  In comparison to \DFM\cite{tewari2023diffusion}, our approach leverages strong image- and video-priors to achieve noticeably higher image fidelity.

\begin{figure*}

\begin{center}
\includegraphics[width=0.88\textwidth, trim={0cm 18cm 0.cm 0cm},clip, page=1]{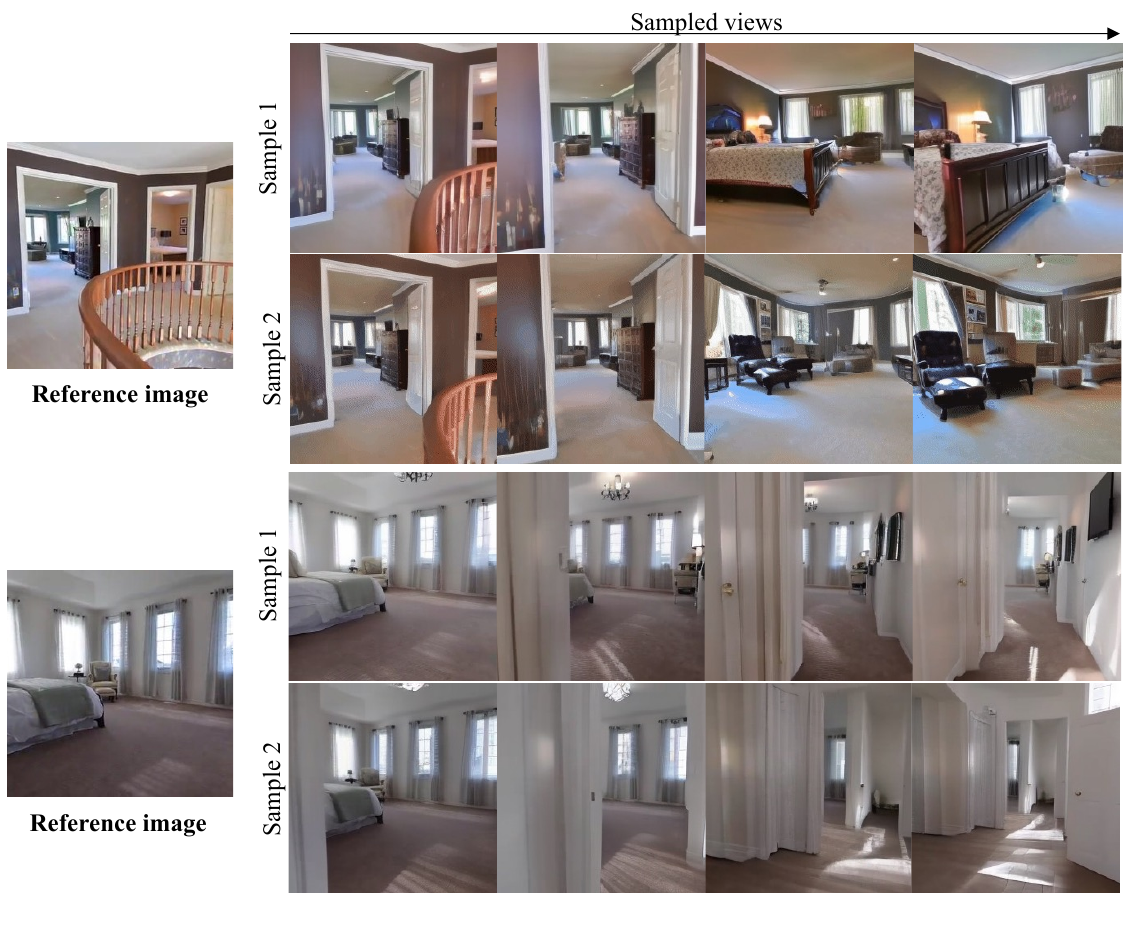}

\end{center}
\vspace{-1.5em}
\caption{
Different samples generated by our probabilistic approach using the same reference image and target trajectory. 
}
\label{fig:add_stochasticy}

\end{figure*}

Furthermore, we demonstrate the stochasticity of our approach in \cref{fig:add_stochasticy} where using the same reference image and target poses, our probabilistic method synthesizes multiple plausible novel views.   

\begin{figure}[h]
\vspace{-3mm}
\begin{center}
\includegraphics[width=.99\columnwidth, page=1, trim={0.3cm 2.5cm 10.4cm 0cm},clip]{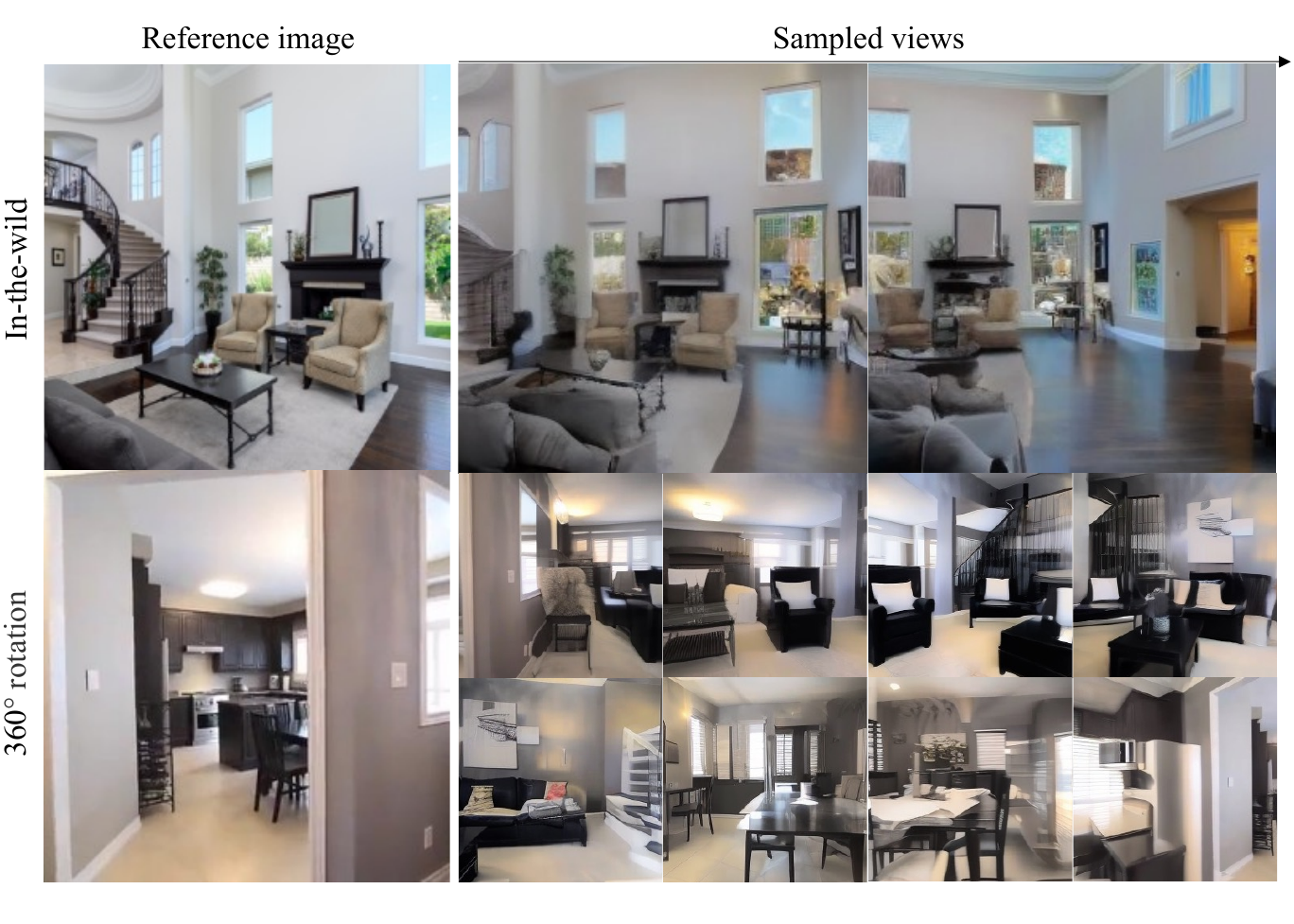}
\end{center}
\vspace{-9mm}
\caption{
Out-of-distribution examples: In-the-wild image with unknown camera parameters and 360° camera rotation.
}
\label{fig:ood}
\vspace{-2mm}
\end{figure}

We also present an in-the-wild and 360° trajectory in \cref{fig:ood}. 

\section{Implementation details}

\subsection{Baselines}
\paragraph{\DFM} 
We use the official implementation of the authors (\url{https://github.com/ayushtewari/DFM.git}). On Realestate10K, we evaluate the provided pre-trained checkpoint. On ScanNet, we train a model from scratch, following the official instructions for RealEstate10k. More specifically, we first train the model at resolution $64\times 64$ for 75K iterations with a total batch size of $16$ on $8$ NVIDIA A100-SXM4-80GB GPUs. Next, we fine-tune the model at resolution $128\times 128$ for 60K iterations using a total batch size of $8$. At both resolutions, larger batch sizes did not fit in the 80GB memory of the GPUs. 
\paragraph{\PhotoNVS} 
We use the official implementation of the authors \url{https://github.com/YorkUCVIL/Photoconsistent-NVS.git} and the provided checkpoint on Realestate10K. On ScanNet, we use the pre-trained VQGAN provided by the authors and train the model at $256 \times 256$ resolution for 500K iterations using $8$ NVIDIA A100-SXM4-80B GPUs with an effective batch size of $64$.
\paragraph{\MVD}
Since \MVD is a purely text-conditional model, we adapt the official implementation (\url{https://github.com/Tangshitao/MVDiffusion.git}) to accept a reference image at inference time. We encode the reference image into latent space and then encode it into the diffusion model’s Gaussian prior space using DDIM inversion. During sampling, the encoded reference image is added to the batch. Since \MVD uses attention layers that operate on all images in the batch jointly, the reference frame affects the sampling for all images. However, during sampling the score estimate is calculated using the full batch, while for DDIM inversion we can only obtain the score estimate for the reference image. In practice, sampling does hence not reproduce the reference image faithfully. We address this issue by additionally optimizing the reference latent after each denoising step to match the reference image. For the optimization at each sampling step, we use Adam~\cite{kingma2015adam} with a learning rate of $0.1$ and train with an L2-Loss and a perceptual loss for 10 iterations.
\paragraph{\TR}
We use the official implementation of the authors (\url{https://github.com/lukasHoel/text2room}) which also supports image-conditional generation. We follow the original setup and use IronDepth~\cite{Bae2022} for depth prediction and StableDiffusion2 inpainting (\url{https://huggingface.co/stabilityai/stable-diffusion-2-inpainting}) for image inpainting.
Since Text2Room formulates the problem as pure depth-to-image/inpainting task, the same pretrained checkpoints can be used for both datasets, RealEstate10K and ScanNet, and no additional training is required. 
\subsection{\Ours}

\paragraph{Training details}

For the encoding of the warped reference images, we use the encoder layers of the pre-trained text-to-image model Stable Diffusion 1.5~\cite{rombach2021highresolution} and use the provided VQ-VAE for latent encoding and decoding.
We initialize the denoising layers of our U-Net model with the pre-trained weights of \VC~\cite{videocrafter}, a latent video diffusion model trained on large scale video data~\cite{Bain21}.
The temporal attention layers serve as strong prior for consistency - see performance of "\Ours no vid." in Table 3 of the main paper and ~\cref{fig:add_abl} for a qualitative comparison. Nevertheless, we fine-tune all layers of the U-Net for the novel view synthesis task to enable the attention layers to learn correspondences between multiple views.
For training, we use Adam~\cite{kingma2015adam} with a learning rate of 1e-05 and batch size of $6$ with $16$ target views per batch at a resolution of $256 \times 256$. Using $8$ NVIDIA A100-SXM4-80B GPUs with an effective batch size of $48$, we train for 300K iterations.
We use DDPM~\cite{song2020denoising} noise scheduling using $t=1000$ time steps for denoising and perform evaluation using DDIM\cite{song2020denoising} sampling with $35$ steps.  

For noise warping, we found that using nearest-neighbor with a receptive field size of 4px at 256px resolution gave the best results. This limited receptive range ensures that the noise distribution remains roughly normal, preventing strong zooms from resulting in a few pixels covering large image portions. 
\paragraph{Inference details}
Using the estimated depth maps with nearest-neighbor interpolation, we calculated the average warping overlap of the initial image with the last frame in the sequence: 20.4\% (24.7\%) on RealEstate10K (ScanNet). The described refinement is applied on poses where the warping overlap is below 20\%, which occurs in 51.8\% (47.1\%) of cases.

\section{Evaluation}
\paragraph{Data processing}
On RealEstate10K, we randomly select 1K sequences with at least 200 frames. For evaluation, for each sequence, we choose a random starting frame at least 200 frames ahead of the last frame. We select 16 frames for evaluation that we uniformly distribute within the interval of 200 views from the starting frame. Following previous evaluation protocols, for short-term evaluation, we set the 5th view to be 50 frames after the starting view in the original video. For long-term evaluation, the last view corresponds to the 200th frame after the starting frame.

On ScanNet, for each of the $100$ test scenes, we sample $10$ starting views ensuring at least 100 frames offset from the last frame in the recordings, resulting in a set of 1K test sequences. 
Since the camera movement in ScanNet recordings is considerably higher and the frame rate noticeably slower compared to RealEstate10K, we consider sequence lengths reduced by 50$\%$ in the original video. 
Therefore, we consider the 25th frame for short-term and the 100th view for long-term evaluation relative to the starting frame.

\begin{figure*}
\begin{center}
\includegraphics[width=0.99\textwidth, trim={0cm 8cm 9.5cm 0.0},clip, page=1]{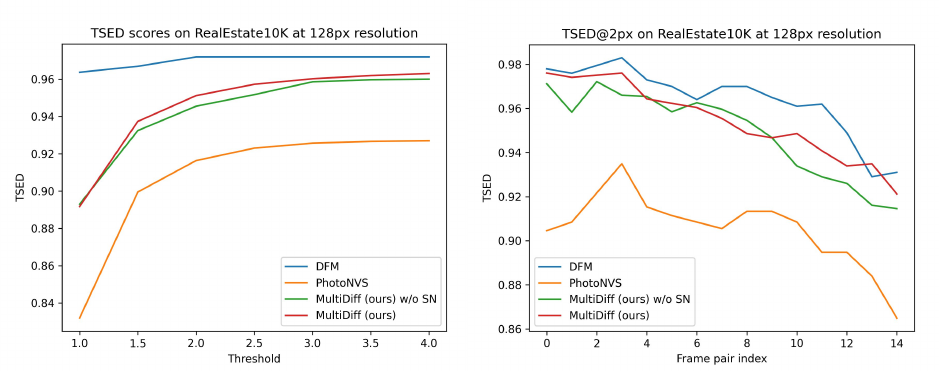}
\end{center}
\vspace{-1.5em}
\caption{
TSED scores evaluated on RealEstate10K at a resolution of $128 \times 128$. The left chart shows the TSED evaluated at different thresholds, on the right we plot the TSED scores over the pairs of frame indices along the trajectory.
}
\label{fig:add_tsed_curves_re10k}

\end{figure*}

\begin{figure*}
\begin{center}
\includegraphics[width=0.99\textwidth, trim={0cm 8cm 9.5cm 0.0},clip, page=2]{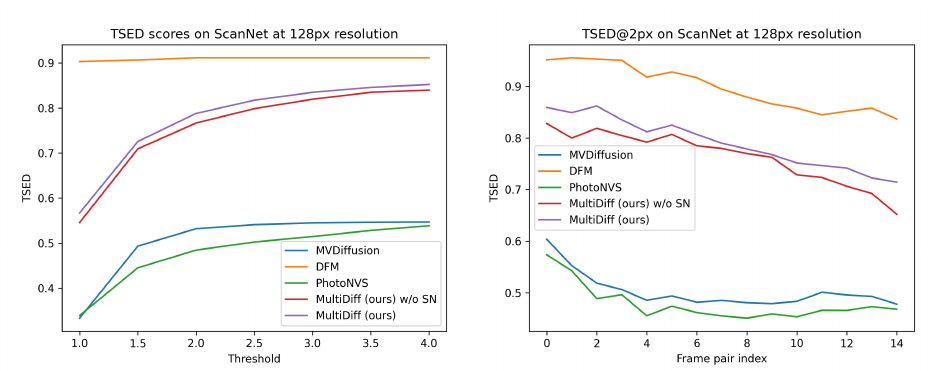}
\end{center}
\vspace{-1.5em}
\caption{
TSED scores evaluated on ScanNet at a resolution of $128 \times 128$. On the left, we show the TSED evaluated at different thresholds. The right chart plots the TSED scores over the pairs of frame indices along the trajectory.
}
\label{fig:add_tsed_curves_scannet}

\end{figure*}

\paragraph{TSED}To compute TSED scores~\cite{Yu2023PhotoconsistentNVS}, we use the official implementation from (\url{https://github.com/YorkUCVIL/Photoconsistent-NVS}) and provide additional quantitative results in~\cref{fig:add_tsed_curves_scannet} and ~\cref{fig:add_tsed_curves_re10k}. While \DFM achieves the highest consistency scores due to its PixelNeRF~\cite{yu2021pixelnerf} formulation, it suffers from noticeably worse image generation quality compared to \Ours (see Table 1 and 2 of the main paper as well as ~\cref{fig:add_re10k_comp} and ~\cref{fig:add_scannet_comp}). As \DFM does not support higher image resolutions, we measure TSED at $128\times 128$ resolution. 

\paragraph{Inference speed}
We report inference performances of \PhotoNVS, \DFM, and \Ours in \cref{tab:benchmark}. As our approach infers multiple frames in parallel and uses an efficient attention architecture, we observe noticeably shorter inference times while achieving higher image fidelity and consistency than the baselines. We note that our approach also scales to larger resolutions as the underlying latent video prior can easily be tuned for image sizes. This is in stark contrast to baselines like \PhotoNVS, \DFM for which the computational costs quickly become too high and require infeasible amounts of memory when trained on larger resolutions.

\paragraph{}
\begin{table}
    \centering
    \resizebox{.40\columnwidth}{!}{
    \begin{tabular}{c|c}
           128px & s/frame\\
         \hline
         \PhotoNVS & 45.6 \\
         \DFM & 17.4 \\
         \Ours & 1.02 \\
    \end{tabular}}
    \resizebox{.40\columnwidth}{!}{
    \begin{tabular}{c|c}
           256px & s/frame\\
         \hline
         \PhotoNVS & 183 \\
         \DFM & - \\
         \Ours & 1.94 \\
    \end{tabular}}
    \caption{Comparison of the inference speed evaluated in seconds per frame using FP32 on an NVIDIA A100-SXM4-80GB. By jointly inferring multiple frames in parallel and using efficient attention architecture, we achieve noticeably shorter inference times.} 
    \label{tab:benchmark}
      \vspace{-2mm}
\end{table}

\begin{figure}[H]
\begin{center}
\includegraphics[width=.99\columnwidth, page=2, trim={0.0cm 2.5cm 10.4cm 0cm},clip]{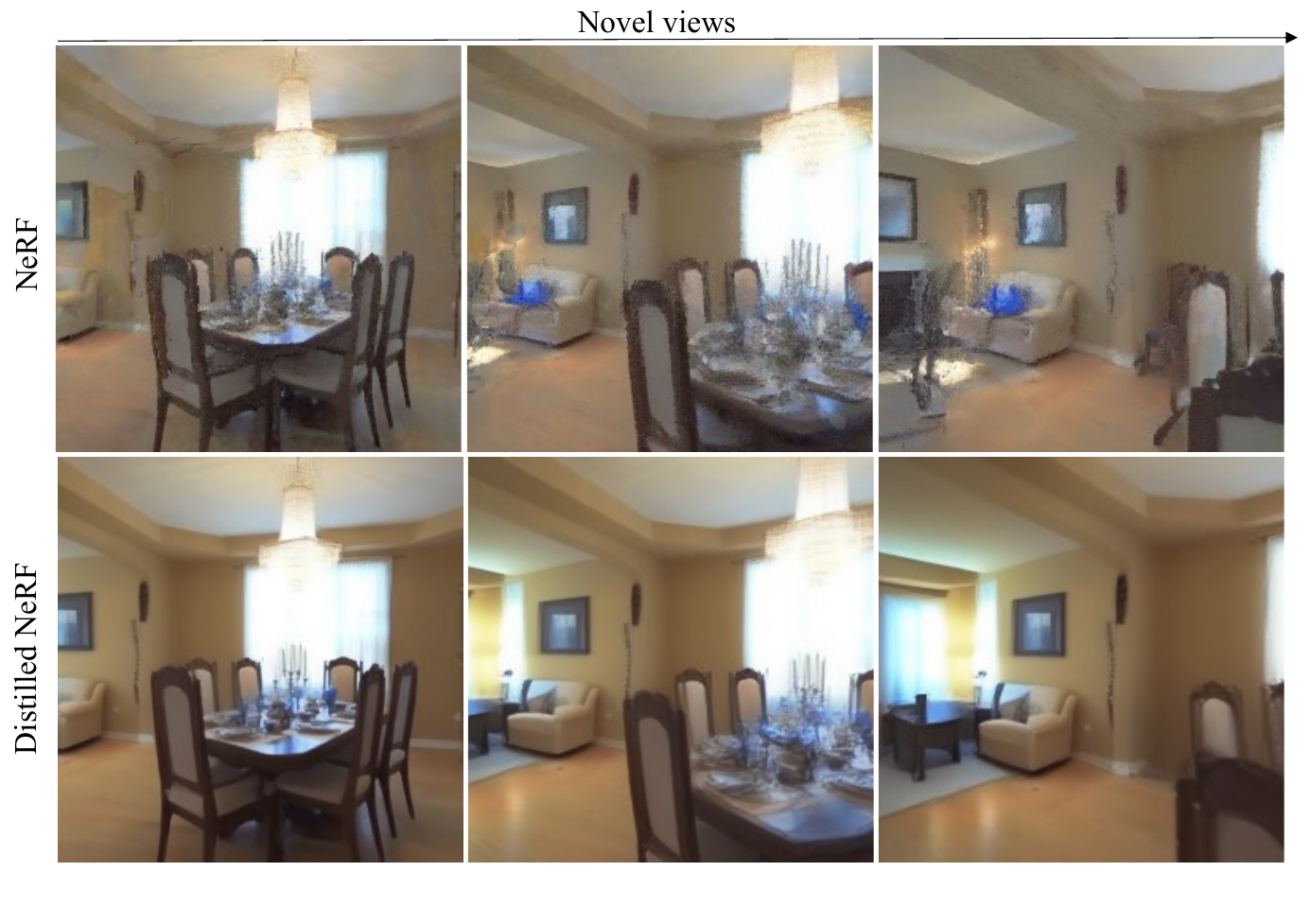}
\end{center}
\vspace{-9mm}
\caption{
NeRF fitting from 16 synthesized views. Row 1: Nerfacto, row 2: Nerfacto + Distillation
}
\label{fig:rebuttal_nerf}
\end{figure}

\paragraph{Fitting a NeRF}
Results for fitting a NeRF with Nerfacto are shown in \cref{fig:rebuttal_nerf}, yielding small pixel-level inconsistencies with floating artifacts. As in \textit{ReconFusion}~\cite{wu2023reconfusion}, we use distillation to obtain a cleaner representation (second row in \cref{fig:rebuttal_nerf}).

\section{Ablations}
\begin{figure*}
\begin{center}
\includegraphics[width=0.88\textwidth, trim={0cm 11cm 0.0cm 0cm},clip, page=1]{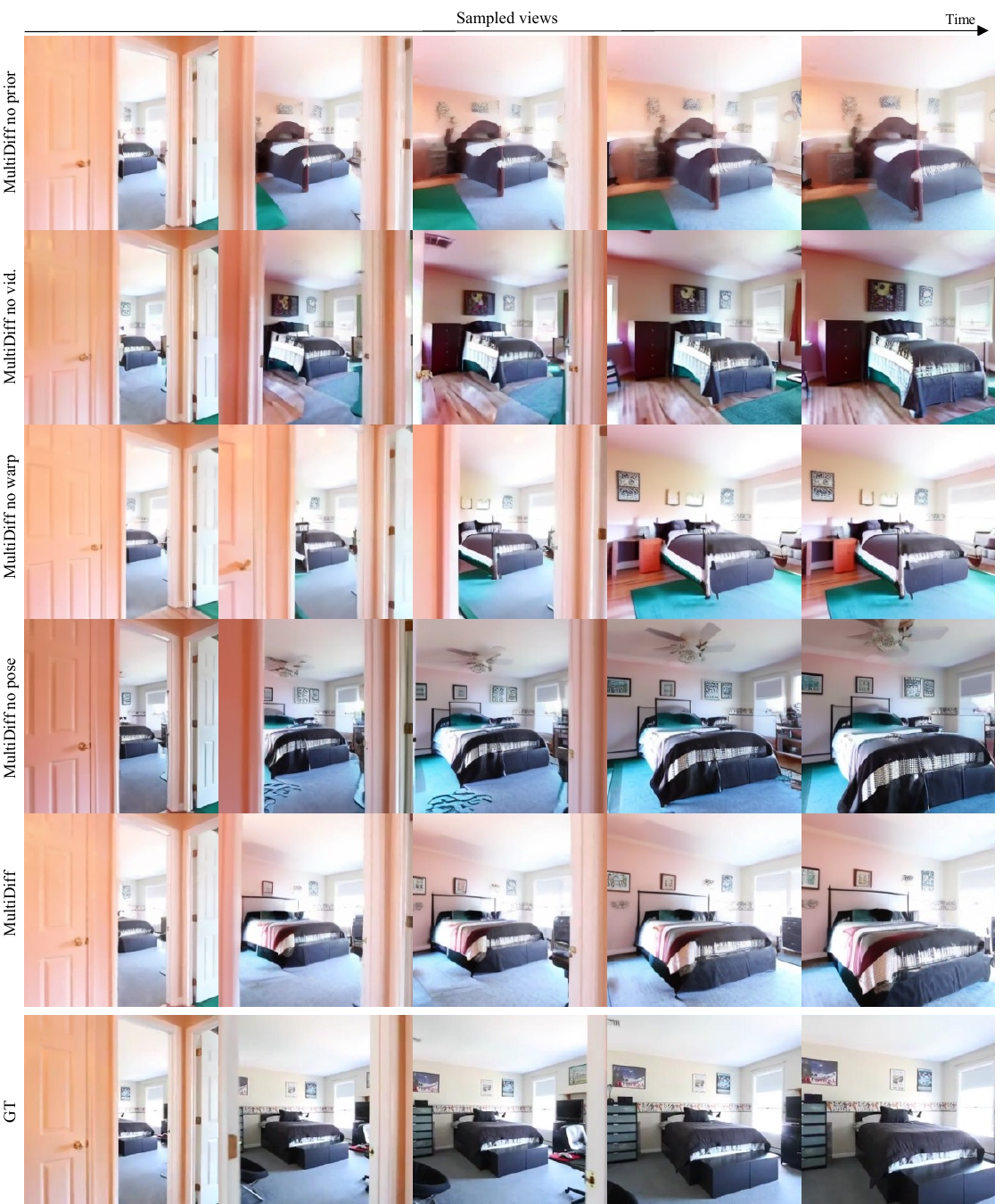}

\end{center}
\vspace{-1.5em}
\caption{
Qualitative comparison of the different ablations of our method on a RealEstate10K test sequence. 
}
\label{fig:add_abl}

\end{figure*}

We show additional qualitative results ablating our design decisions in~\cref{fig:add_abl}. We note that training our model from scratch ("\Ours no prior") leads to over-smoothed results that do not closely follow the target trajectory. Furthermore, we showcase the effect of using the image prior but not initializing the weights of our correspondence attention layer with the weights of the pre-trained video prior (see "\Ours no vid."): The results are overall less consistent as e.g. the floor changes from carpet to wood.
Our method uses depth-based image warps to reproject the reference image to the target poses, providing strong cues about the target views. We ablate the importance of this in Table 3 of the main table (see "\Ours no warp") and show an example of in ~\cref{fig:add_abl}. Without using the warps of the reference image, our model is not able to faithfully follow the target trajectory.
As under strong camera motion, there is little to no overlap with the reference image, we also learn an embedding of the target pose and show the effect of removing this information ("\Ours no pose") in ~\cref{fig:add_abl}. Using the additional pose embeddings provides additional guidance about the target poses leading to better TSED scores.

\begin{table}
    \centering
    \resizebox{.99\columnwidth}{!}{
    \begin{tabular}{c|c|cccc|cccc}
           \multirow{2}{*}{Datset} & \multirow{2}{*}{Method} & \multicolumn{4}{c|}{Short-term} & \multicolumn{4}{c}{Long-term} \\
          & & PSNR $\uparrow$ & LPIPS ↓ & FID ↓  & KID ↓ & FID ↓  & KID ↓ & FVD  ↓ & mTSED  $\uparrow$  \\
         \hline
         \multirow{2}{*}{RE10K}
         & \Ours + IronDepth & 15.49  &0.402 & 28.36  & 0.005 & 34.91 & 0.006 & 115.72 &  0.797  \\
         & \Ours + ZoeDepth & \textbf{15.65} & \textbf{0.393} & \textbf{25.90} & \textbf{0.004} & \textbf{30.15} & \textbf{0.006} &  \textbf{105.9} &  \textbf{0.855} \\
         \midrule
         \multirow{2}{*}{ScanNet}
          & \Ours + IronDepth & \textbf{15.05} & 0.435 & 44.15 & 0.010 & \textbf{46.87} & 0.013 &  118.3  &    0.503 \\
         & \Ours + ZoeDepth &  15.00 & \textbf{0.431} & \textbf{43.84}  & \textbf{0.010}  & 47.11  &  \textbf{0.013}  & \textbf{114.9}  &  \textbf{0.576}  \\

    \end{tabular}}
    \caption{Qualitative comparison of using IronDepth trained on ScanNetv2 as alternative depth estimator evaluated $256\times 256$ resolution. We notice that the results are comparable for the short-term metrics. For long-term evaluation, we observe that the non-metric scaling of IronDepth leads to worse mTSED scores.} 
    \label{tab:iron}
      \vspace{-2mm}
\end{table}

In addition, we ablate the effect of using an alternative depth estimator to ZoeDepth~\cite{ZoeDepth} in~\cref{tab:iron}. For this, we use  IronDepth~\cite{Bae2022} pretrained on ScanNetv2 and report qualitative results on RealEstate10K and ScanNet. While we observe comparable results in image quality performance, we note that using IronDepth leads to worse consistency scores. As IronDepth does not provide estimates in metric scale, using these depth estimates to warp the reference image leads to less accurate conditional information. Ultimately, this results in lower consistency scores - see e.g., $mTSED$ that decreases by $\approx 6\%$ on RealEstate10K and $\approx 13\%$ on ScanNet using IronDepth compared to using ZoeDepth.

\begin{figure}[H]
\begin{center}
\includegraphics[width=1.0\columnwidth, trim={0.4cm 3.cm 9.9cm 0cm},clip, page=1]{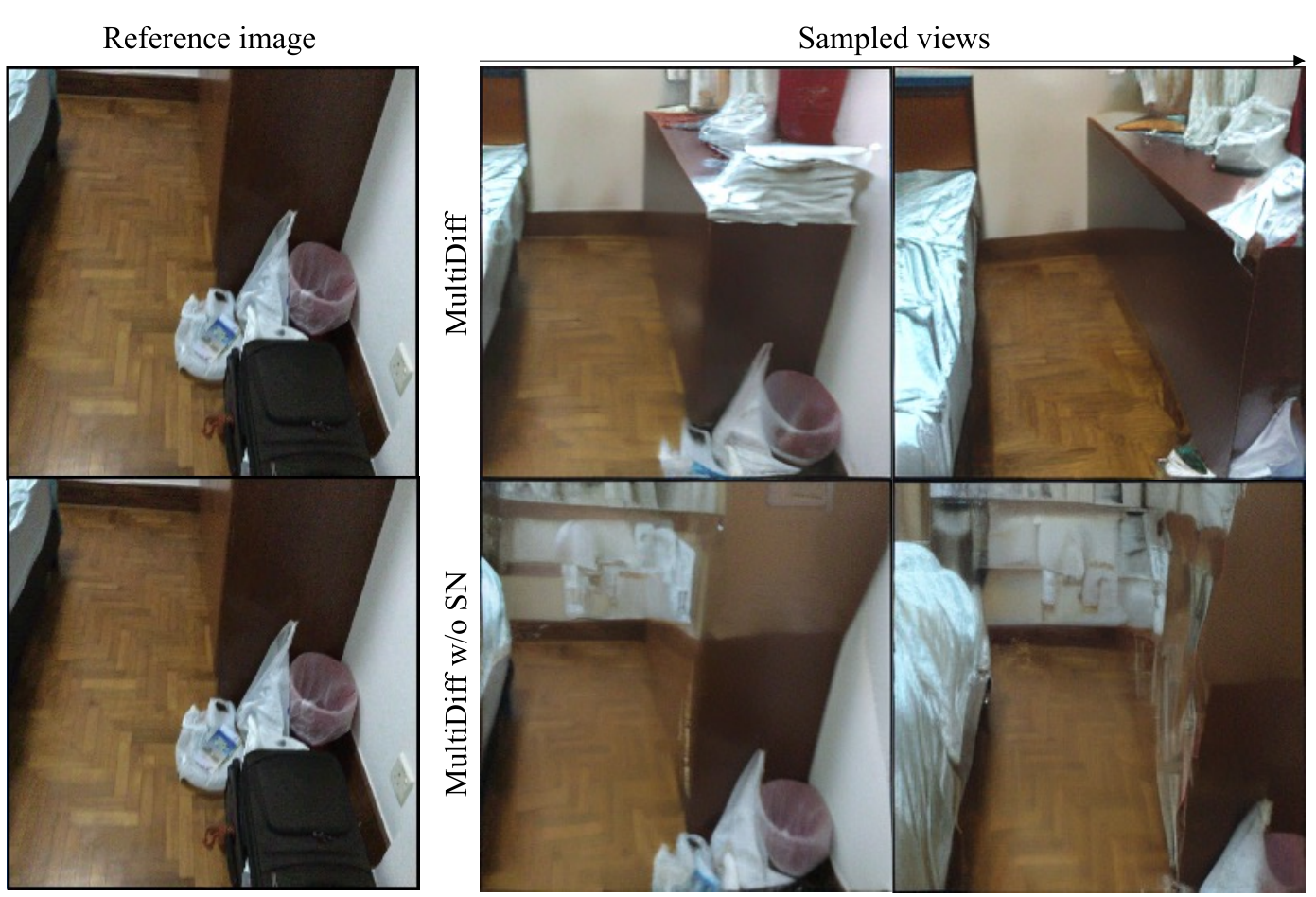}

\end{center}
\vspace{-1.5em}
\caption{
Additional qualitative comparison of applying structured noise on a ScanNet test sequence. Applying structured noise leads to more consistent and overall more realistic sampling results.
}
\label{fig:add_struct_noise}

\end{figure}

Furthermore, we qualitatively show the effect of using structured noise in~\cref{fig:add_struct_noise} on a ScanNet test sequence. We note that by structuring the noise using the depth estimates, we obtain more realistic and consistent synthesis results.

\end{document}